%% file: collas2026_conference.tex
\documentclass{article} %
\usepackage{collas2026_conference,times}
\usepackage{easyReview}
\usepackage{booktabs}
\usepackage{amssymb}
\usepackage{subcaption} 
\usepackage{algorithm, algorithmic}
\usepackage{wrapfig}
\usepackage{listings}

\usepackage[symbol]{footmisc}

\input{math_commands.tex}

\usepackage{hyperref}
\hypersetup{
    colorlinks=true,
    linkcolor=red,
    filecolor=magenta,
    urlcolor=blue,
    citecolor=purple,
    pdftitle={Overleaf Example},
    pdfpagemode=FullScreen,
    }
\usepackage[dvipsnames]{xcolor}

\title{Learning to Forget: Continual Learning with Adaptive Weight Decay}

\author{Aditya A. Ramesh$^{1}$, Alex Lewandowski$^{2}$, J\"{u}rgen Schmidhuber$^{3, 1}$ \\
$^{1}$The Swiss AI Lab, IDSIA USI-SUPSI, Lugano, Switzerland\\
$^{2}$University of Alberta, Alberta Machine Intelligence Institute, Edmonton, Canada\\
$^{3}$Center of Excellence for Generative AI, KAUST, Thuwal, Saudi Arabia\\
\texttt{\{aditya, juergen\}@idsia.ch, lewandowski@ualberta.ca}
}

\preprintcopy %

\begin{document}

\maketitle

\begin{abstract}
Continual learning agents with finite capacity must balance acquiring new knowledge with retaining the old.
This requires controlled forgetting of knowledge that is no longer needed, freeing up capacity to learn.
Weight decay, viewed as a mechanism for forgetting, can serve this role by gradually discarding information stored in the weights.
However, a fixed scalar weight decay drives this forgetting uniformly over time and uniformly across all parameters, even when some encode stable knowledge while others track rapidly changing targets. 
We introduce Forgetting through Adaptive Decay (FADE), which adapts per-parameter weight decay rates online via approximate meta-gradient descent.
We derive FADE for the online linear setting and apply it to the final layer of neural networks.
Our empirical analysis shows that FADE automatically discovers distinct decay rates for different parameters, complements step-size adaptation, and consistently improves over fixed weight decay across online tracking and streaming classification problems.
\end{abstract}

\section{Introduction}

Judicious forgetting is essential for continual learning. 
An agent with finite capacity cannot retain everything that was previously learned.
Therefore, some degree of forgetting in a controlled way~\citep{french1999catastrophic, kumar2025continual} is necessary to successfully navigate the stability-plasticity trade-off \citep{grossberg1987competitive, elsayed2024addressing}. 
For short-term memories stored in activations, gating mechanisms in recurrent neural networks provide learned forgetting \citep{Hochreiter:97lstm, gers2000learning}. 
But there is no corresponding mechanism for long-term knowledge stored in slowly changing weights.

Weight decay (e.g.,~\citealp{hanson1988comparing,krogh1991simple}) is widely used as a regularizer in deep learning~\citep{Loshchilov2017DecoupledWD}, biasing each weight toward zero at every update.
In typical stationary settings, training data is revisited multiple times and weight decay primarily facilitates regularization~\citep{mackay1992practical}.
Here, we focus on the online non-stationary setting, where data arrives one sample at a time, past samples are never revisited, and task boundaries are unknown. 
In such settings, weight decay can serve as a forgetting mechanism that controls how much past information each weight retains.

We explore the complementary interpretation of weight decay as a mechanism for forgetting.
One limitation of weight decay as a forgetting mechanism is that the decay rate is a single scalar, fixed across all parameters and constant over time.
While several works have explored adaptive weight decay for mini-batch training \citep{ishii2017layer, nakamura2019adaptive, xie2023overlooked}, these methods focus on stationary settings with fixed datasets.
In continual learning, a fixed decay rate, even with appropriate tuning, is fundamentally mismatched to the problem. 
Some parameters may encode stable knowledge that should be retained, while others track rapidly changing targets that require fast forgetting. 
An effective forgetting mechanism must adapt selectively, assigning each parameter its own rate.

To this end, we introduce \textit{\textbf{F}orgetting through \textbf{A}daptive \textbf{DE}cay (FADE)}, which adapts per-parameter weight decay rates via gradient-based meta-learning.
FADE uses meta-gradients (e.g.~\citealp{hochreiter2001learning, xu2018meta}) that approximate forward-mode differentiation (e.g.,~\citealp{robinson1987utility, williams1989learning}), enabling online updates.
Specifically, FADE derives from the same meta-gradient approximations as IDBD~\citep{sutton1992adapting, degris2024step}, which adapts per-parameter step sizes (learning rates) online.
FADE instead focuses on how much to forget.

We derive FADE for the online linear setting and apply it to the final layer of neural networks. 
On a linear tracking problem, FADE discovers distinct decay rates for relevant and irrelevant features and complements per-parameter step-size adaptation via IDBD. 
On a nonlinear teacher-student tracking problem, FADE with SGD achieves roughly half the error of AdamW~\citep{Loshchilov2017DecoupledWD}. 
On streaming label-permuted EMNIST, FADE outperforms weight clipping, the best prior method~\citep{elsayed2024weight}. 
Across all settings, FADE is robust to the initialization of the decay rate, recovering strong performance even from poor initializations.

\section{Forgetting through Adaptive Decay (FADE)}
\label{method}

\begin{algorithm}[bt]
\caption{FADE: Forgetting through Adaptive Decay (online linear regression)}
\label{alg:fade}
\begin{algorithmic}[1]
\REQUIRE step size $\alpha$, meta-step size $\theta_\lambda$, initial decay parameter $\gamma_0 \in \mathbb{R}^d$
\STATE Initialize weights $w_0 \in \mathbb{R}^d$, traces $g_0 \leftarrow \mathbf{0} \in \mathbb{R}^d$
\STATE Initialize $\lambda_0^i \leftarrow \exp({\gamma_0^i})$ for all $i$
\FOR{$t = 0, 1, 2, \ldots$}
    \STATE Receive input $x_t \in \mathbb{R}^d$ and target $y^*_t \in \mathbb{R}$
    \STATE Predict $y_t \leftarrow \langle w_t, x_t \rangle$
    \STATE Compute error $\delta_t \leftarrow y^*_t - y_t$
    \FOR{each parameter $i = 1, \ldots, d$}
        \STATE \textit{\# Adapt decay rate}
        \STATE $\gamma_{t+1}^i \leftarrow \gamma_t^i + \theta_\lambda \, \delta_t \, x_t^i \, g_t^i$
        \STATE $\lambda_{t+1}^i \leftarrow \exp({\gamma_{t+1}^i})$
        \STATE \textit{\# Update sensitivity trace}
        \STATE $g_{t+1}^i \leftarrow g_t^i \big[1 - \lambda_{t+1}^i - \alpha (x_t^i)^2 \big]^{+} - \lambda_{t+1}^i \, w_t^i$
        \STATE \textit{\# Update weight with adaptive decay}
        \STATE $w_{t+1}^i \leftarrow (1 - \lambda_{t+1}^i) \, w_t^i + \alpha \, \delta_t \, x_t^i$
    \ENDFOR
\ENDFOR
\end{algorithmic}
\end{algorithm}

FADE replaces a static, global weight decay hyperparameter with dynamic, per-parameter decay adapted online via meta-gradients. 
In this section, we derive FADE for the online linear regression setting using the same forward-mode differentiation techniques and approximations as IDBD~\citep{sutton1992adapting}. 
The key idea is to parameterize the decay rate ($\lambda_i)$ for each parameter $w^i$ as $\lambda^i = \exp({\gamma^i})$ and update the meta-parameter $\gamma^i$
by gradient descent on the prediction error. 
Since the effect of $\gamma^i$
on the error is mediated through the weight update, this requires tracking the sensitivity $\partial w^i / \partial \gamma^i$ via an auxiliary trace $g^i$, which is maintained online.

\paragraph{FADE for online linear regression.} At every time step $t \in \{0, 1, 2, \dots \}$, the learner receives features $x_t \in \mathbb{R}^d$ and target $y_t^* \in \mathbb{R}$.
The learner maintains weights $w_t \in \mathbb{R}^d$ that are used for predicting $y_t = \langle w_t, x_t \rangle$.
Let $\delta_t = y^*_t - y_t$ be the error, and $J_t = \delta_t^2/2$ be the loss at time step $t$.

We update weights with weight decay~\citep{hanson1988comparing} and the delta-rule, equivalent to SGD for linear regression with squared error~\citep{widrow1960adaptive},
\begin{equation}
\label{eqn:weight_update_wd_delta_rule}
    w_{t+1} = (1 - \lambda_{t+1}) \, w_{t} + \alpha \, \delta_t \, x_t,
\end{equation} where $\alpha$ is a scalar step size and $\lambda_{t+1} \in \mathbb{R}^d$ controls per-parameter decay, with $\lambda_{t+1}^i$ controlling the decay for the $i$-th component of the weight $w_t^i$.
Note that unrolling the weight update yields an exponentially weighted sum with an effective memory horizon of $\sim 1/\lambda^i$. 
FADE adapts this horizon per-parameter.
Increasing $\lambda^i$ shortens the horizon (faster forgetting), decreasing it lengthens the horizon (more retention).

The weight decay coefficients $\lambda_t$ are parameterized using $\gamma_t \in \mathbb{R}^d$. Specifically, $\lambda_{t}^i = \exp (\gamma_t^i), \ i \in \{ 1, 2, \dots d\} $.
The meta-parameters $\gamma$ are updated via gradient descent by differentiating the loss with respect to $\gamma$ through the weight update (Equation~\ref{eqn:weight_update_wd_delta_rule}).
This can be understood as an online form of cross-validation where the weights are updated on the current sample, and the resulting performance on the next sample provides gradients for the meta-parameters~\citep{xu2018meta}.

The meta-gradients are approximated using forward-mode differentiation~\citep{robinson1987utility, williams1989learning}  with the assumption that changing $\gamma^i$ primarily affects $w^i$, with negligible effect on other weights. 
Full derivations are provided in Appendix~\ref{app:derivations}.

Concretely, FADE uses the following updates:
\begin{align}
\label{eq:fade}
    \gamma^i_{t+1} &\leftarrow \gamma^i_t + \theta_\lambda \, \delta_t \, x^i_t \, g^i_t, \quad \lambda^i_{t+1} \leftarrow \exp({\gamma^i_{t+1}}),\\
        g^i_{t+1} &\leftarrow g^i_t \, [1 - \lambda^i_{t+1} - \alpha (x^i_t)^2]^+ - \lambda^i_{t+1} \, w_t^i\\ 
    w_{t+1}^i &\leftarrow (1 - \lambda_{t+1}^i) \, w_t^i + \alpha \, \delta_t \, x^i_t.
\end{align}

Here $g^i$ is a trace tracking $\partial w^i / \partial \gamma^i$, the sensitivity of the weight to its decay meta-parameter, and $\theta_\lambda$
is the meta step size. 
The $[\cdot]^+$ denotes $\max(\cdot, 0)$, a positive-bounding 
operation for stability.
See Algorithm~\ref{alg:fade} for the pseudocode.

FADE can be combined with strategies that use adaptive coordinate-wise step sizes, like Adam~\citep{kingma2014adam} or IDBD~\citep{sutton1992adapting}. 
In particular, when combined with IDBD, both the decay rate and step size are adapted per-parameter via meta-gradients. 
We provide the derivation and pseudocode for FADE with IDBD in Appendix~\ref{app:derivations:fade_idbd} and Algorithm~\ref{alg:fade-idbd}, respectively.

\paragraph{Computational cost.} 
FADE adds two scalar states per parameter ($\gamma^i$ and $g^i$) 
and one scalar hyperparameter ($\theta_\lambda$), preserving $\mathcal{O}(d)$ cost per step of online gradient descent.

\paragraph{FADE with neural networks.} FADE is derived for the online linear setting, where the prediction is a linear function of the weights. 
In a neural network, the final layer is a linear function of the hidden representation, so FADE can be applied directly to the final layer while using any standard optimizer (SGD, Adam) for the hidden layers.
This approach of applying a meta-gradient method only to the final layer has also been used successfully by \citet{javed2024swifttd}.
We follow this approach in our nonlinear experiments (Sections~\ref{exps:nonlinear} and~\ref{exps:emnist}).
The cross-entropy extension, which replaces the squared-error gradient with the softmax gradient, is derived in Appendix~\ref{app:derivations:fade_ce}.

\section{Experiments}
\label{exps}

We evaluate FADE on three online non-stationary problems of 
increasing complexity\footnote[2]{Code is available at \url{https://github.com/Aditya-Ramesh-10/Fade}.}
.
First, we consider a linear tracking task where FADE's derivation applies exactly (Section~\ref{exps:linear}).
Next, we investigate a nonlinear teacher-student tracking task where FADE is applied 
to the final layer (Section~\ref{exps:nonlinear}).
Finally, we consider a streaming classification benchmark in which we apply FADE with the cross-entropy loss to the final layer (Section~\ref{exps:emnist}). 
Across all three experiments, we examine whether adaptive decay improves performance and robustness to hyperparameter choice compared to fixed decay.

\subsection{Linear Tracking}
\label{exps:linear}

\begin{wrapfigure}{r}{0.33\columnwidth}
    \centering
\includegraphics[width=0.3\columnwidth]{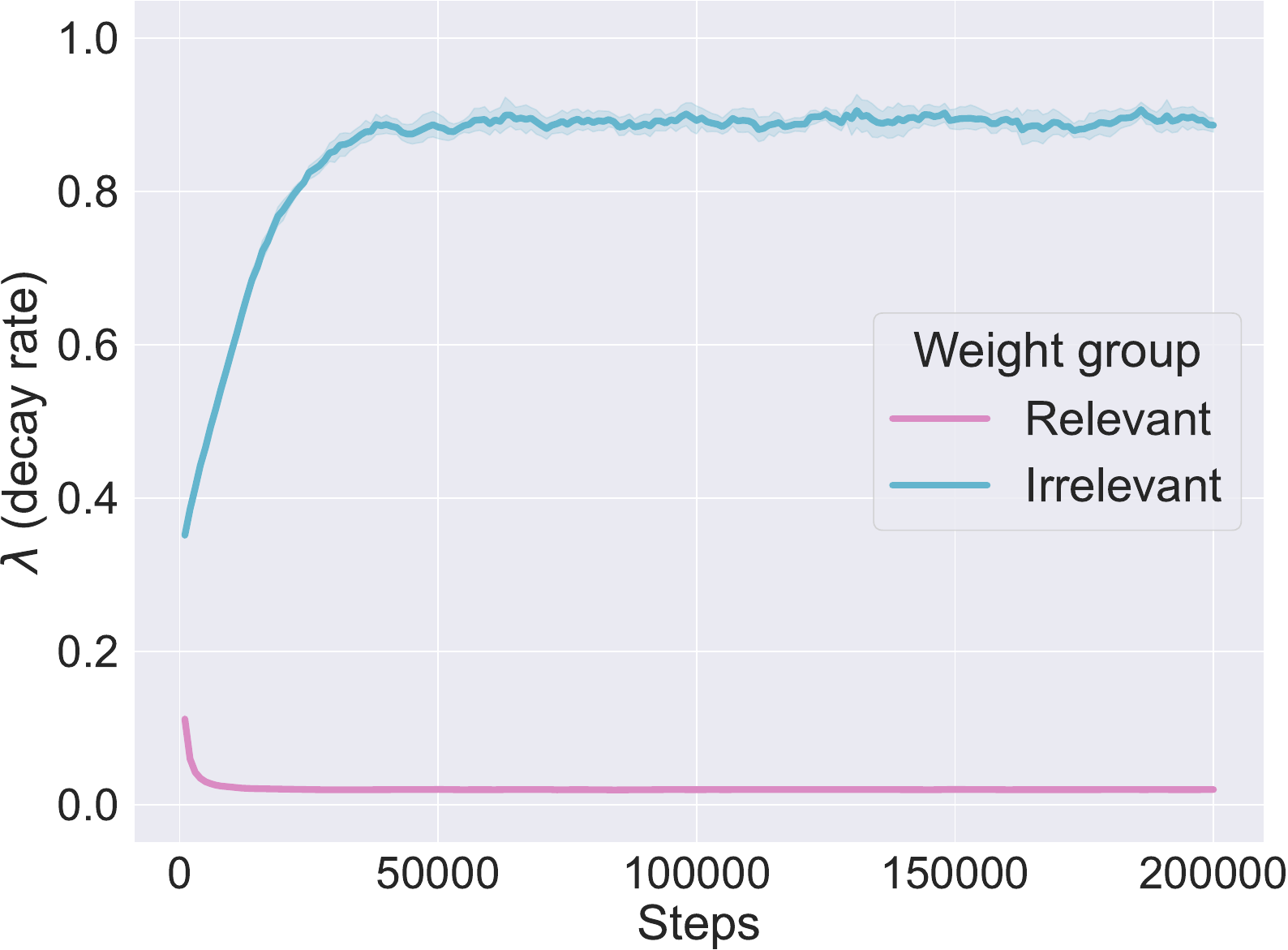}
    \caption{Evolution of FADE's average decay rates for relevant and irrelevant weight groups on the linear tracking problem with zero noise, starting from $\lambda_0 \approx 0.3
(\gamma_0 = -1.2)$, with $\alpha = 0.1$
and $\theta_\lambda = 0.01$.
    }
\label{fig:linear_lambda}
\vspace{-30pt}
\end{wrapfigure}
\paragraph{Setup.}
We consider a non-stationary tracking task that was previously used to motivate step size adaptation with meta-gradients~\citep{degris2024step}.
The learner is presented with a $d=20$ dimensional input at every time step, sampled i.i.d.\ $x_t \in \mathbb{R}^d \sim \mathcal{N}(0, I_d)$, and target $y_t^* = \mathbf{w}^*_t \cdot x_t + \epsilon$, where $\epsilon \sim \mathcal{N}(0, \sigma_n^2)$.
We consider two values of $\sigma_n$.
Of the 20 weights of $\mathbf{w}^*_t$, 5 are relevant (either $\pm 1$) and 
15 are irrelevant (fixed at 0). 
Every 20 steps, one randomly chosen relevant weight flips sign.

\paragraph{Evaluation.}
We compare stochastic gradient descent (SGD, equivalent to the delta rule), IDBD 
\citep{sutton1992adapting} (adaptive step sizes), and FADE 
(adaptive decay with SGD), along with fixed weight decay (WD) variants. 
In the variant with adaptive step sizes and weight decay (FADE + IDBD), we tie the meta-step sizes $\theta_{\lambda} = \theta_{\alpha} = \theta$.
We report MSE over all 200K steps of interaction for the best 
hyperparameters per method, averaged across 10 seeds.
The hyperparameter search configuration is provided in Appendix~\ref{app:impdets:linear_tracking}.

\paragraph{Results.}
Results are presented in Table~\ref{tab:linear_tracking}. 
SGD uses a single step size for all weights and incurs the largest error. 
Adding fixed-weight decay improves SGD by biasing unused weights toward zero.
IDBD adapts step sizes per parameter, gradually driving those for irrelevant inputs toward zero, and substantially outperforms SGD. 
Fixed weight decay also benefits IDBD, reducing lifetime MSE from $1.486$ to $1.301$ in the noiseless setting.
FADE, which adapts decay rates rather than step sizes, improves considerably over SGD + WD ($1.653$ vs.\ $2.726$).
FADE discovers distinct decay rates. 
Weights associated with irrelevant features converge to $\lambda \approx 0.9$, decaying toward zero.
Weights for relevant features settle near $\lambda \approx 0.02$, maintaining a longer memory (Figure~\ref{fig:linear_lambda}).

As this problem was designed to highlight step-size adaptation, IDBD's advantage over FADE here is not particularly surprising. 
The key challenge is distinguishing relevant from irrelevant features, which IDBD handles by learning to stop updating irrelevant weights.
Crucially, FADE + IDBD achieves the lowest error overall ($1.246$), outperforming both FADE and IDBD alone. 
This indicates that adaptive step sizes and adaptive decay are complementary mechanisms.
IDBD controls how quickly each parameter incorporates new information, while FADE controls how much past information each parameter retains.
Even on a problem tailored to step-size adaptation, dynamically adjusting the forgetting rate via FADE provides an additional benefit.

\paragraph{Coupled adaptation of step size and weight decay.}
We also compare against a `coupled' variant of step size and weight decay adaptation where the decay term comes from the $L_2$ regularized loss, i.e., the decay term is $\alpha^i \lambda^i$ rather than $\lambda^i$ alone, and the weight update is $w^i_{t+1} = (1 - \alpha^i_{t+1}\lambda^i_{t+1}) \, w^i_t + \alpha^i_{t+1} \, \delta_t \, x^i_t$  (see Appendix~\ref{app:derivations:coupled} for the derivation).
The best coupled variant achieves an MSE of $1.270 \pm 0.006$ with $\sigma_n=0$ and $2.670\pm0.009$ with $\sigma_n=1$, slightly worse than (decoupled) FADE+IDBD ($1.246 \pm 0.006$ and $2.646 \pm 0.009$).
We observe that the coupled update introduces a shared $-\alpha^i \lambda^i w^i_t$ term into both traces, potentially causing some interference in adaptation.

\begin{table}[t]
\centering
\caption{Average MSE $\pm$ standard deviation across 10 seeds on the linear tracking task across two noise levels $\sigma_n$.}
\label{tab:linear_tracking}
\begin{tabular}{lcc}
\toprule
& $\sigma_n = 0$ & $\sigma_n = 1.0$ \\
\midrule
SGD            & $3.628 \pm 0.021$ & $5.119 \pm 0.026$ \\
SGD + WD       & $2.726 \pm 0.008$ & $4.087 \pm 0.012$ \\
IDBD           & $1.486 \pm 0.007$ & $2.937 \pm 0.011$ \\
IDBD + WD      & $1.301 \pm 0.006$ & $2.718 \pm 0.010$ \\
FADE           & $1.653 \pm 0.009$ & $3.044 \pm 0.011$ \\
\textbf{FADE + IDBD}    & $\mathbf{1.246 \pm 0.006}$ & $\mathbf{2.646 \pm 0.009}$ \\
\bottomrule
\end{tabular}
\end{table}

\subsection{Non-linear tracking}
\label{exps:nonlinear}

\paragraph{Setup.}
Vector targets are obtained from a teacher network.
Concretely, the teacher is a neural network with input dimension $d=32$ and a hidden layer of size $h=256$ with ReLU activations.
The output layer (head) is linear and outputs a vector $y^* \in \mathbb{R}^{20}$.
The learner is a student network that has the same architecture as the teacher. 
At each step, the learner receives a single input $x_t \sim \mathcal{N}(0, I_d)$ and the teacher's output $y^*_t$ as the target. 
Learning proceeds online from a single sample per step, with no replay buffer or resets.

To introduce non-stationarity, we periodically change weights in the teacher's final layer.
Of the $20$ output units, $6$ are stable, and incoming weights remain fixed throughout the interaction. 
A further 7 are fast-changing: every 
$P=500$ steps, each weight into these output units is independently multiplied by a random sign ($\pm 1$ with equal probability).
The remaining 7 output units are slow-changing, undergoing the same random sign perturbation every 
$15P=7500$ steps. 
This creates three distinct rates of non-stationarity across the output units, while the teacher's input-to-hidden weights remain fixed throughout.

\begin{table}[b]
\centering
\caption{Average MSE $\pm$ standard deviation over the final 500K steps across 5 seeds on the non-linear tracking problem.}
\label{tab:nonlinear_tracking}
\begin{tabular}{lc}
\toprule
Method & MSE (final 500K steps) \\
\midrule
         SGD           & $0.0168 \pm 0.0002$ \\
         SGD + WD           & $0.0167 \pm 0.0002$  \\
         Adam & $0.0170 \pm 0.0002$ \\
         AdamW & $0.0138 \pm 0.0001$ \\
         \textbf{FADE + SGD} & $\mathbf{0.0073 \pm 0.0001}$  \\
         FADE + Adam & $0.0087 \pm 0.0001$ \\
\bottomrule
\end{tabular}
\end{table}

\paragraph{Evaluation.}
The learner interacts for 2M steps. 
We report average MSE over the final 500K steps, averaged across 5 seeds. 
Since FADE is derived for the linear setting, we apply it only to the final layer of the student network. 
In FADE+SGD, SGD is used throughout the network. In FADE+Adam, Adam is used throughout, with FADE providing adaptive decay on the head. 
We compare SGD, SGD with weight decay (SGD+WD), Adam (no weight decay), AdamW, FADE+SGD, and FADE+Adam.
For each approach, we perform a grid search over the step-size and weight decay hyperparameters.
Hyperparameter search details and selected hyperparameters are provided in Appendix~\ref{app:impdets:nonlinear_tracking}.

\paragraph{Results.}

\begin{figure}[t]
\begin{subfigure}{.49\textwidth}
  \centering
  \includegraphics[width=0.85\linewidth]{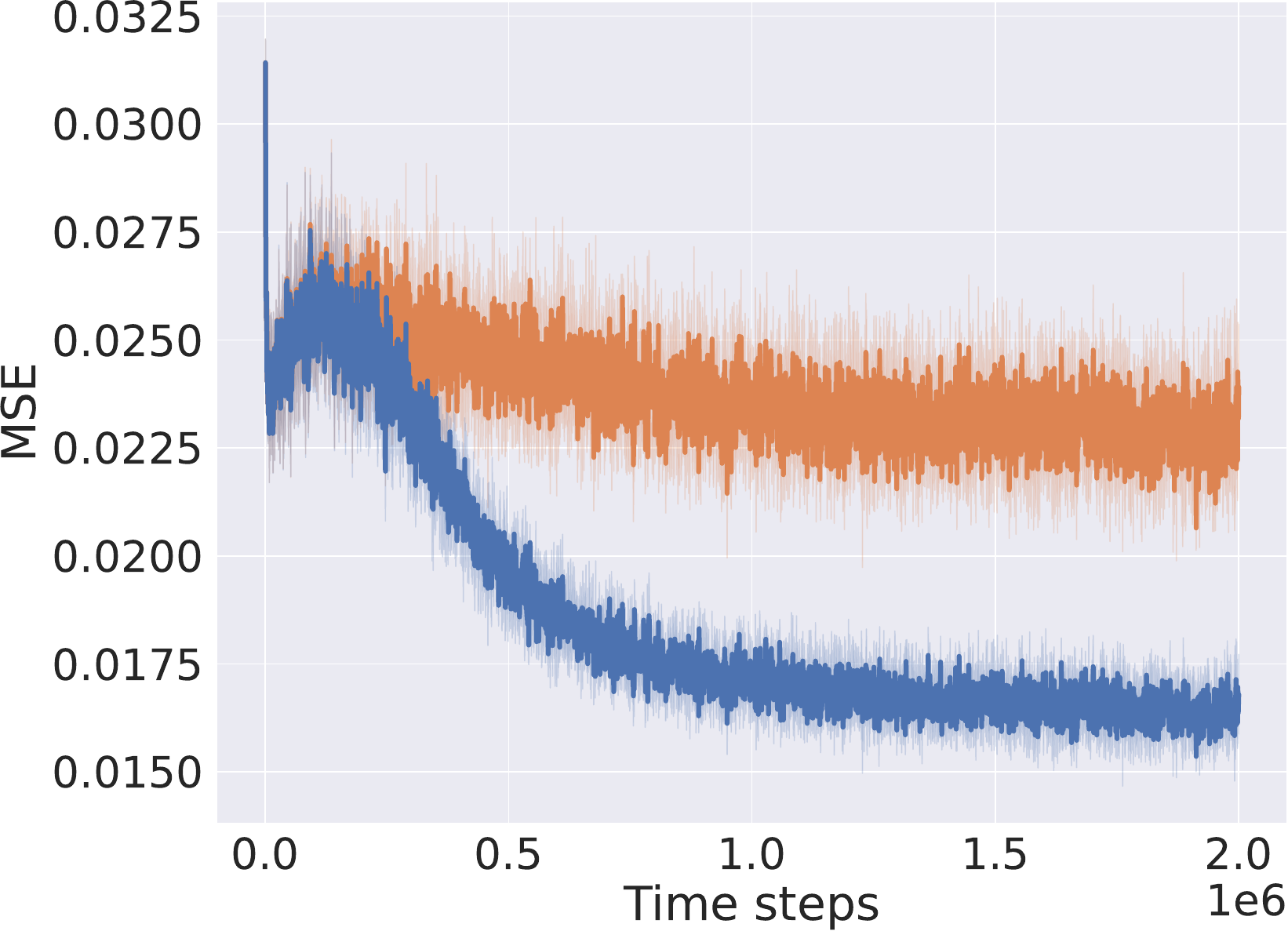}
  \subcaption{Fast}
  \label{fig:nonlinear_tracking_fast_mse}
\end{subfigure}%
\begin{subfigure}{.49\textwidth}
  \centering
  \includegraphics[width=0.85\linewidth]{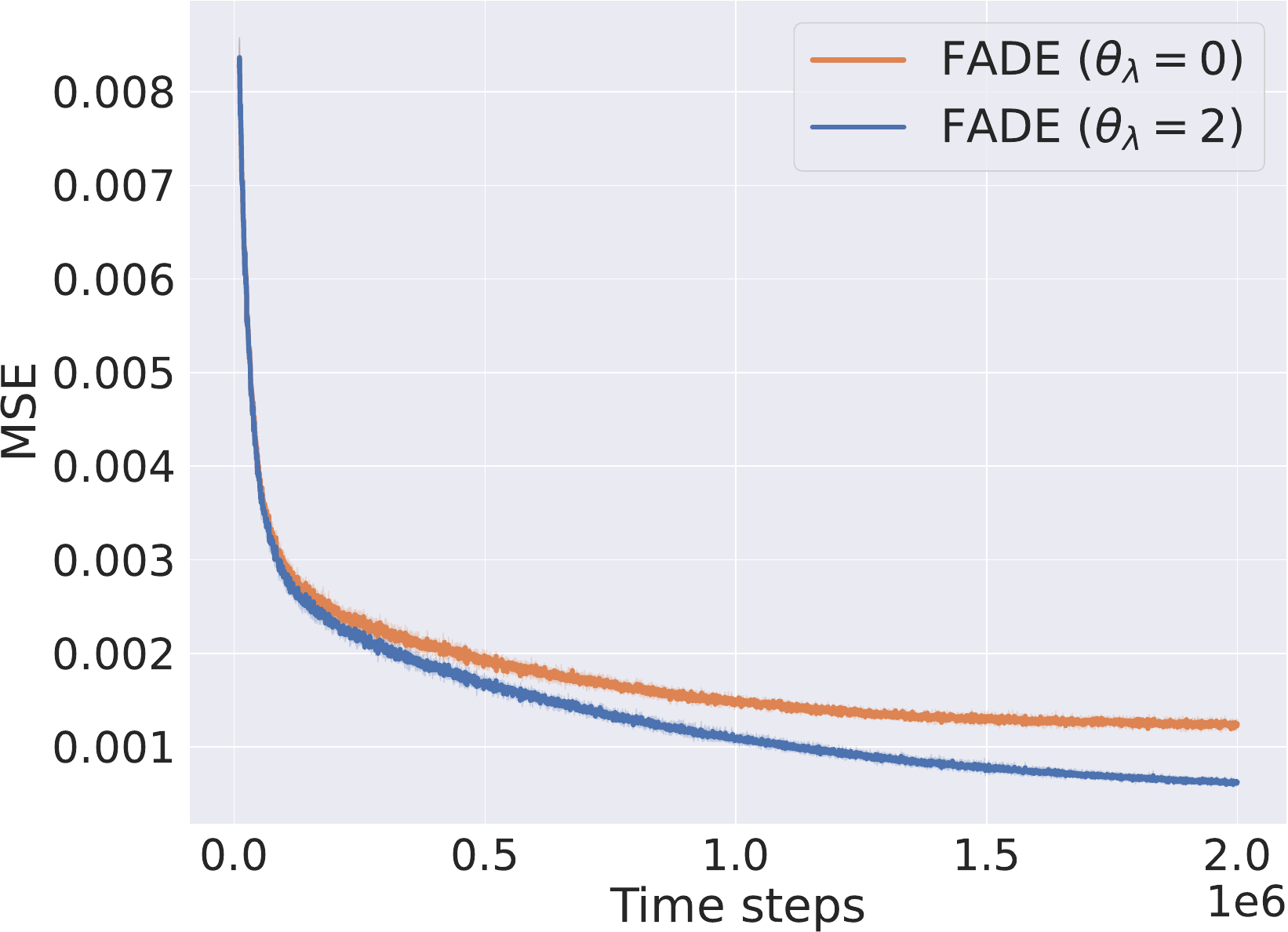}
  \subcaption{Stable}
  \label{fig:nonlinear_tracking_stable_mse}
\end{subfigure}

\vspace{0.5em}

\begin{subfigure}{.49\textwidth}
  \centering
  \includegraphics[width=0.85\linewidth]{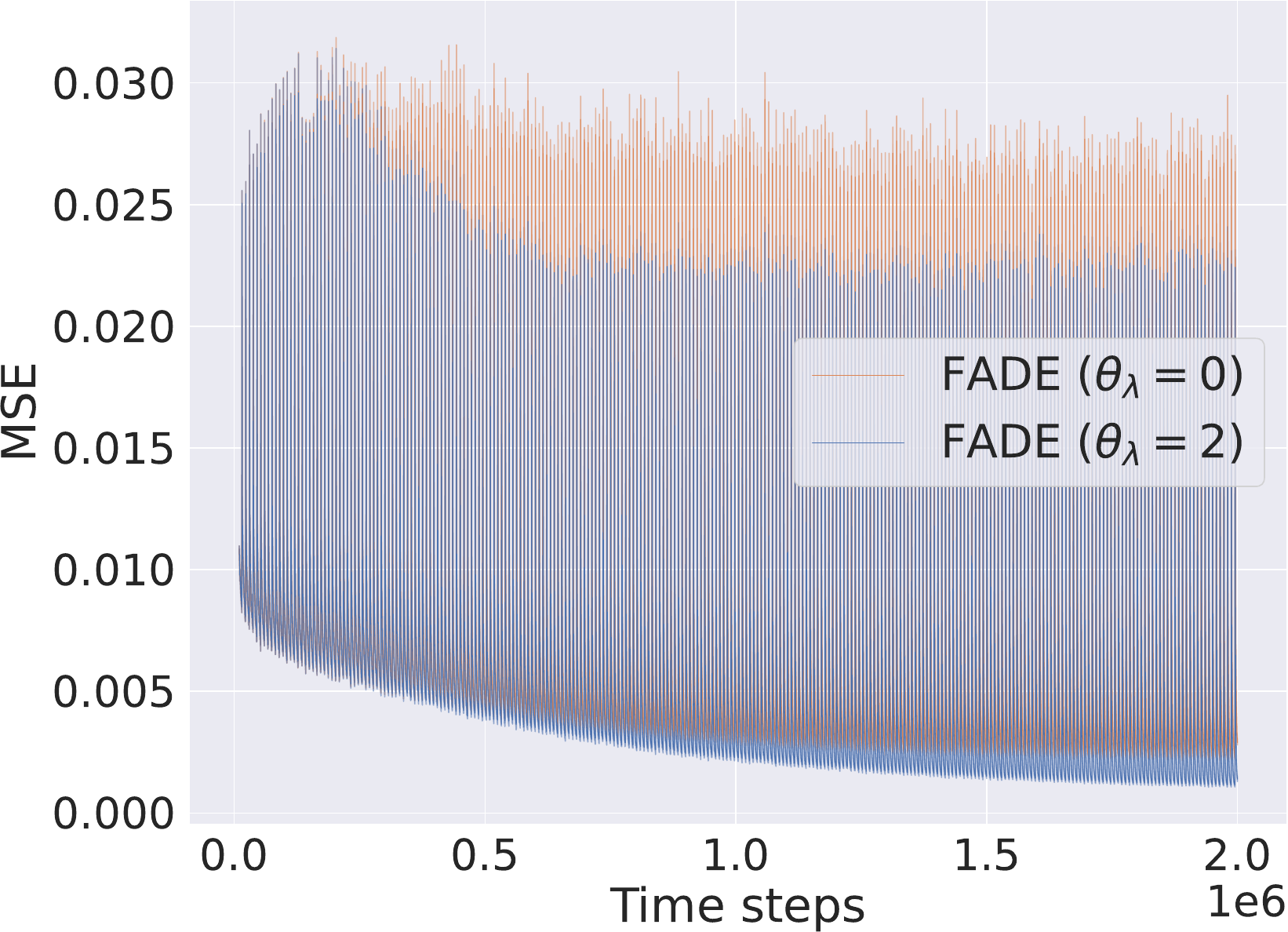}
  \subcaption{Slow}
  \label{fig:nonlinear_tracking_slow_mse}
\end{subfigure}%
\begin{subfigure}{.49\textwidth}
  \centering
  \includegraphics[width=0.85\linewidth]{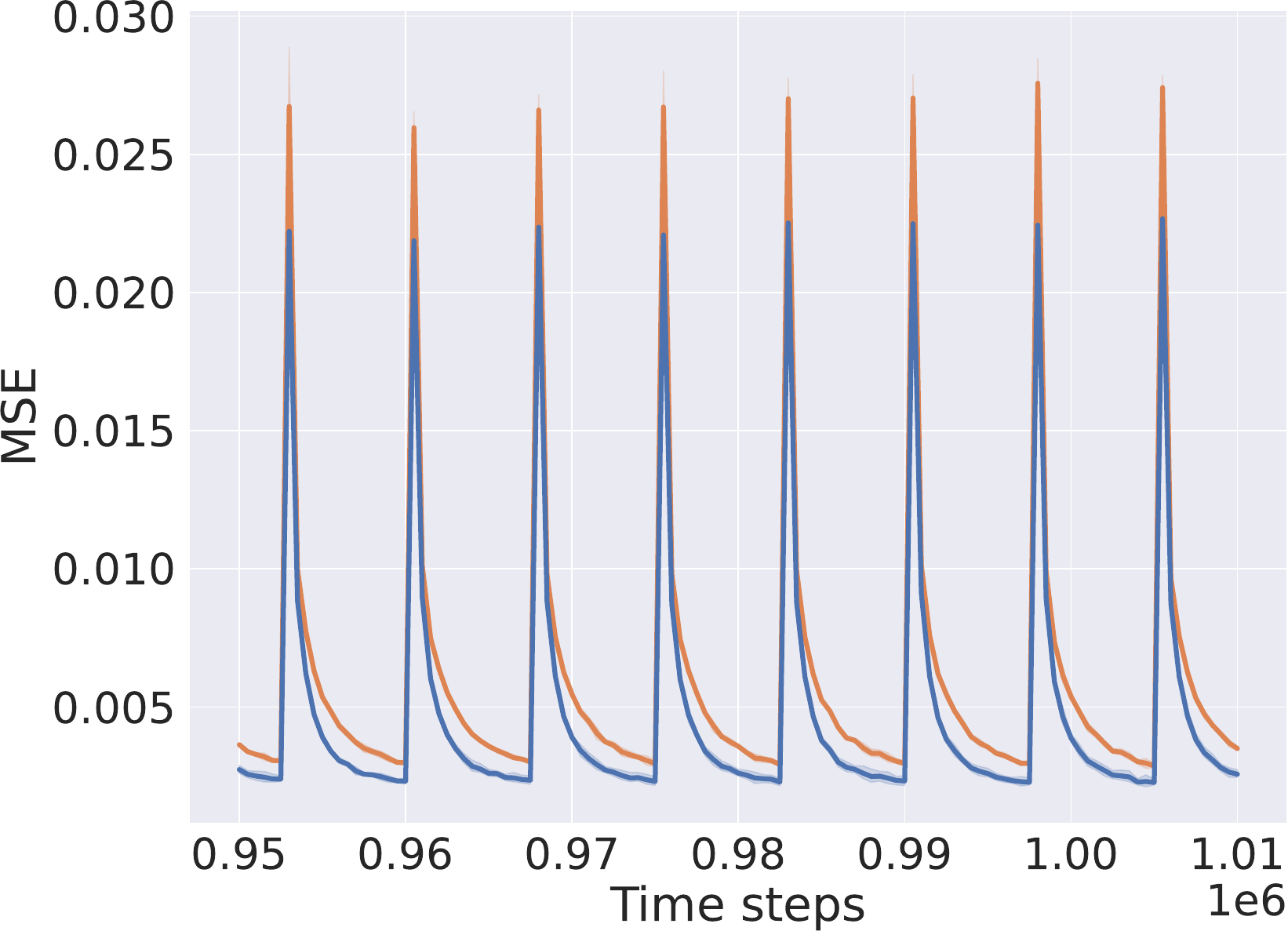}
  \subcaption{Slow, zoomed in}
  \label{fig:nonlinear_tracking_slow_mse_zoomed}
\end{subfigure}
\caption{MSE by group on nonlinear tracking problem. With $\gamma_0 = -9.2 (\lambda_0\approx0.0001)$, comparing FADE+SGD ($\theta_{\lambda}=2$) with its fixed-decay counterpart FADE+SGD ($\theta_{\lambda} =0$).}
\label{fig:nonlinear_tracking_groupwise_mse}
\end{figure}

\begin{wrapfigure}{r}{0.3\columnwidth}
\vspace{-10pt}
    \centering
\includegraphics[width=0.25\columnwidth]{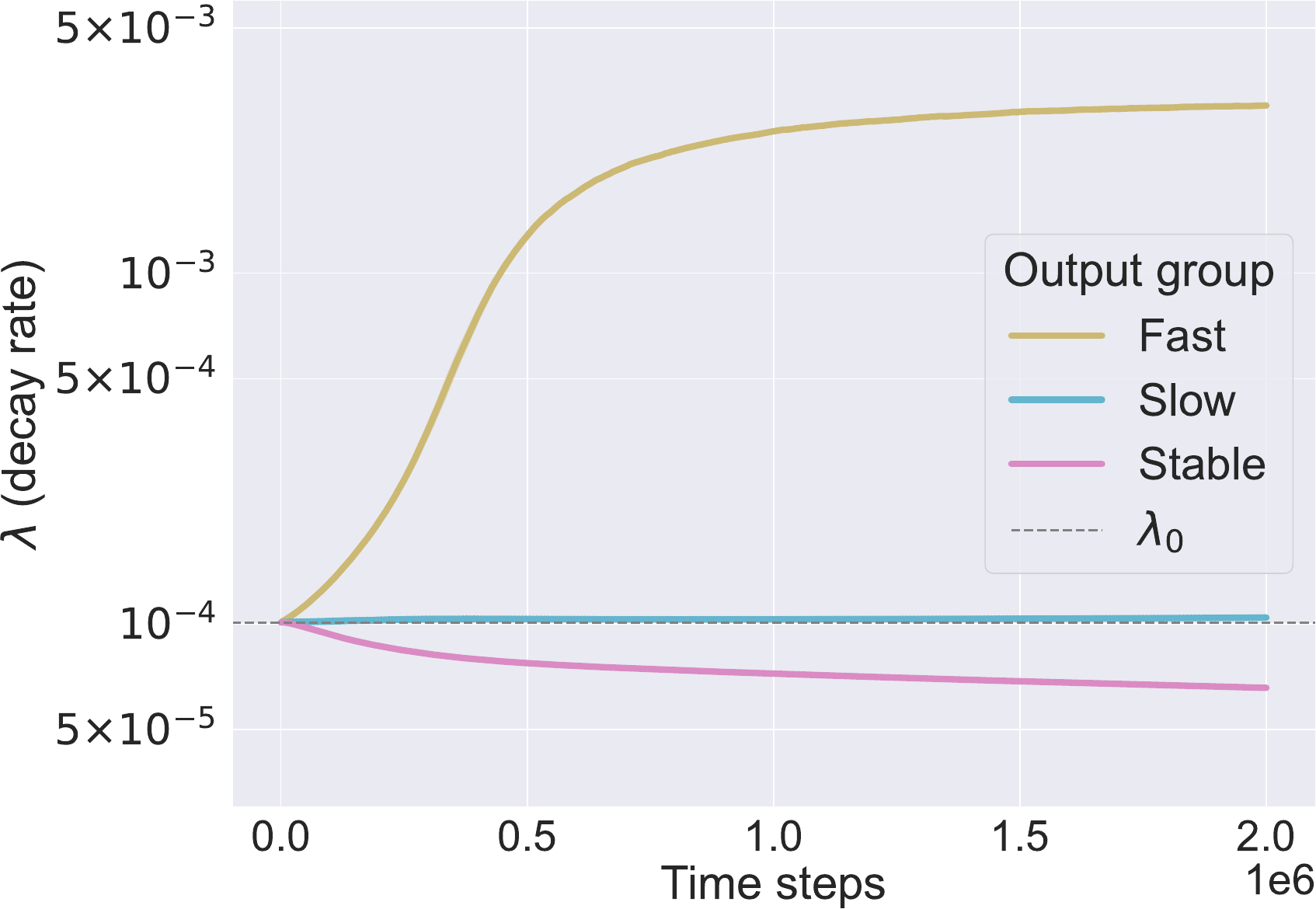}
    \caption{Evolution of FADE's average per-group decay rates on the nonlinear tracking problem starting from a shared initialization $\lambda_0 \approx 10^{-4}$ (dashed line), and $\theta_{\lambda} = 2.0$.
    }
\label{fig:nonlinear_tracking_sgdfade_lambda_evolution}
\vspace{-10pt}
\end{wrapfigure}

Main results are presented in Table~\ref{tab:nonlinear_tracking}.
All baseline methods (SGD, SGD+WD, Adam, AdamW) achieve similar MSE over the final 500K steps, ranging from around $0.014$ to $0.017$.
AdamW is the best performing baseline, achieving an average of $0.0138$.
Using FADE helps both SGD and Adam.
FADE+SGD achieves $0.0073$, roughly half the error of AdamW, while FADE+Adam achieves $0.0087$.
We observe that FADE+SGD outperforms FADE+Adam, likely because FADE's meta-gradient derivation holds exactly for SGD, but only approximately for Adam (Appendix~\ref{app:derivations:fade_adam}).

To isolate FADE's contribution, we compare FADE+SGD with adaptation ($\theta_{\lambda}=2$) against its non-adaptive counterpart ($\theta_{\lambda}=0$).
With $\theta_{\lambda}=0$, FADE reduces to SGD with fixed weight decay on the head and no decay on the hidden layer.
Figure~\ref{fig:nonlinear_tracking_groupwise_mse} shows per-group MSE for $\gamma_0=-9.2$, comparing FADE $(\theta_{\lambda}=2)$ with fixed decay $(\theta_{\lambda}=0)$ . 
Across all three groups (fast, slow, and stable outputs), we see improvements from FADE's adaptation, suggesting that the adaptation is effective for a wide range of non-stationarity rates.
Figure~\ref{fig:nonlinear_tracking_sgdfade_lambda_evolution} shows the corresponding evolution of weight decay across the groups starting from a shared initialization. 
Here, FADE increases decay for fast-changing outputs and decreases it for stable outputs, with slow outputs receiving an intermediate rate.

We examine sensitivity to the initial decay rate $\lambda_0$ and meta-step-size $\theta_{\lambda}$ in Table~\ref{tab:fade_sensitivity}.
Our results show that adaptive weight decay via FADE consistently improves performance.
With fixed decay ($\theta_{\lambda}=0$), performance varies substantially across the decay rate, with the best initialization $(\gamma_0=-6.9)$ achieving $0.0099$ and the worst $(\gamma_0=-2.3)$ achieving $0.0360$.
With FADE $(\theta_{\lambda} > 0)$, the sensitivity-gap narrows considerably. 
Even from the worst initialization $(\gamma_0=-2.3)$, FADE recovers to $0.0125$. The FADE+Adam sensitivity (Table~\ref{tab:fade_adam_sensitivity}) shows a similar pattern, confirming that the benefits are not specific to SGD.

Additionally, we note that (fixed) weight decay on the head, but not on the hidden layer, can perform reasonably well with a well-tuned decay (see $\gamma_0 = -6.9$ with $\theta_{\lambda}=0$ in Table~\ref{tab:fade_sensitivity}).
However, adaptive weight decay via FADE provides benefits in terms of improved MSE and robustness to the choice of the initial decay.

\begin{table}[b]
\centering
\caption{FADE + SGD sensitivity to initial decay rate $\gamma_0$ and meta-step-size $\theta_\lambda$ on the non-linear tracking problem. $\theta_\lambda=0$ corresponds to fixed weight decay on the head. 
Results are average MSE $\pm$ standard deviation over the final 500K steps across 5 seeds.}
\label{tab:fade_sensitivity}
\begin{tabular}{lcccc}
\toprule
 & $\theta_\lambda=0$ & $\theta_\lambda=0.5$ & $\theta_\lambda=1.0$ & $\theta_\lambda=2.0$ \\
\midrule
$\gamma_0=-2.3\ (\lambda_0\approx0.1)$   & $0.0360 \pm 0.0011$   & $0.0148 \pm 0.0001$  & $0.0137 \pm 0.0001$ &  $0.0125 \pm 0.0001$\\
$\gamma_0=-4.6\ (\lambda_0\approx0.01)$  & $0.0182 \pm 0.0003$  & $0.0098 \pm 0.0001$  & $0.0093 \pm 0.0001$  & $0.0089 \pm 0.0001$ \\
$\gamma_0=-6.9\ (\lambda_0\approx0.001)$ & $0.0099 \pm 0.0001$ & $0.0084 \pm 0.0001$ & $0.0082 \pm 0.0001$ & $0.0080 \pm 0.0001$  \\
$\gamma_0=-9.2\ (\lambda_0\approx0.0001)$ & $0.0104 \pm 0.0002$ & $0.0083 \pm 0.0001$ & $0.0075 \pm 0.0001$  & $\mathbf{0.0073 \pm 0.0001}$  \\
\bottomrule
\end{tabular}
\end{table}

\begin{table}[t]
\centering
\caption{FADE + Adam sensitivity to initial decay rate $\gamma_0$ and meta-step-size $\theta_\lambda$ on the non-linear tracking problem. $\theta_\lambda=0$ corresponds to fixed weight decay on the head. 
Results are average MSE $\pm$ standard deviation over the final 500K steps across 5 seeds.
}
\label{tab:fade_adam_sensitivity}
\begin{tabular}{lcccc}
\toprule
 & $\theta_\lambda=0$ & $\theta_\lambda=0.5$ & $\theta_\lambda=1.0$ & $\theta_\lambda=2.0$ \\
\midrule
$\gamma_0=-2.3\ (\lambda_0\approx0.1)$   & $0.0315 \pm 0.0008$   & $0.0126 \pm 0.0001$  & $0.0121 \pm 0.0001$ &  $0.0118 \pm 0.0001$\\
$\gamma_0=-4.6\ (\lambda_0\approx0.01)$  & $0.0179 \pm 0.0002$  & $0.0107 \pm 0.0001$  & $0.0103 \pm 0.0001$  & $0.0102 \pm 0.0001$ \\
$\gamma_0=-6.9\ (\lambda_0\approx0.001)$ & $0.0109 \pm 0.0001$ & $0.0097 \pm 0.0001$ & $0.0097 \pm 0.0001$ & $0.0097 \pm 0.0001$  \\
$\gamma_0=-9.2\ (\lambda_0\approx0.0001)$ & $0.0128 \pm 0.0002$ & $0.0094 \pm 0.0001$ & $\mathbf{0.0087 \pm 0.0001}$  & $0.0088 \pm 0.0001$  \\
\bottomrule
\end{tabular}
\end{table}

\subsection{Streaming image classification with label permutation}
\label{exps:emnist}

\paragraph{Setup}
Following the setup of \citet{elsayed2024weight}, we evaluate on a streaming label-permuted classification problem using the Extended MNIST dataset (EMNIST;~\citealp{cohen2017emnist}), consisting of 47 classes.
At each step, the learner receives a single image, makes a prediction, observes the label, and then performs one gradient update.
The prediction is evaluated before the update to measure online performance.
Every 2500 steps, a fresh random permutation is applied to all class labels, changing the target function.
Since the images remain unchanged across tasks, useful features can, in principle, be retained, but the mapping from features to labels must be relearned after each permutation.
The network is a two-hidden-layer MLP (300, 150) with LeakyReLU activations.

\paragraph{Evaluation} 
We train for 5M steps (2000 label permutation tasks), which is $5$ times the interaction length considered by \citet{elsayed2024addressing}.
We report average online accuracy across the full run.
All methods use their best hyperparameters selected by grid search (see Appendix~\ref{app:impdets:label_perm}).

We compare SGD (with weight decay), Adam, AdamW, and FADE.
Additionally, we compare against SGD with weight clipping (SGD + WClip).
Weight clipping (with SGD) was the best-performing approach on this problem in the experiment conducted by \citet{elsayed2024weight}, outperforming other strategies such as L2 init~\citep{kumar2025maintaining}, Shrink\&Perturb~\citep{ash2020warm}, etc.
For FADE, we again apply it to the final layer that produces the classification logits.
See Appendix~\ref{app:derivations:fade_ce} for FADE with the cross-entropy loss.

\begin{figure}[t]
    \centering
    \begin{minipage}{0.48\columnwidth}
        \centering
        \includegraphics[width=\linewidth]{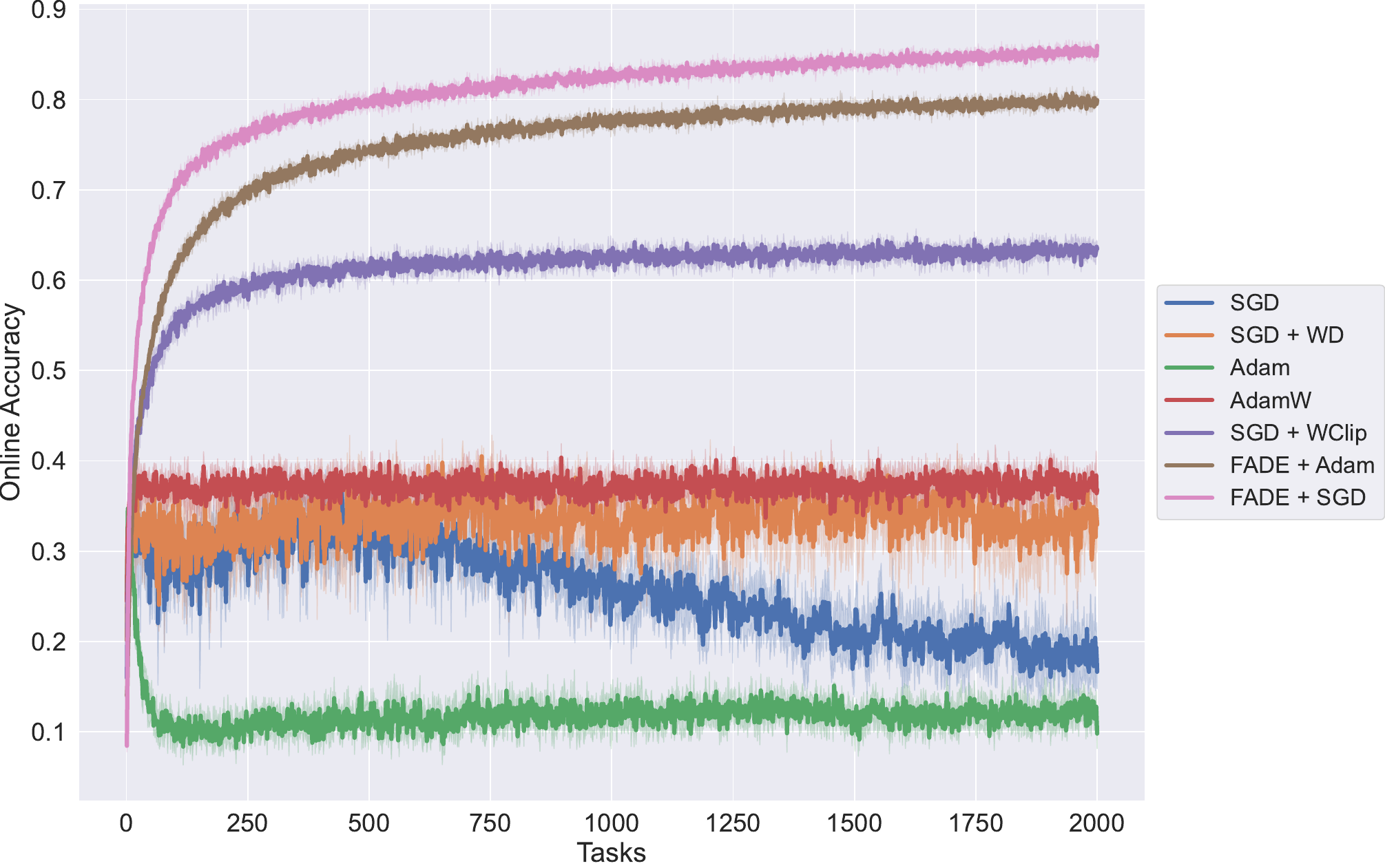}
        \caption{Online accuracy on label-permuted EMNIST.
        FADE+SGD and FADE+Adam apply FADE to the final layer.}
        \label{fig:emnist_lp_full_learning_curves}
    \end{minipage}\hfill
    \begin{minipage}{0.48\columnwidth}
        \centering
        \includegraphics[width=0.9\linewidth]{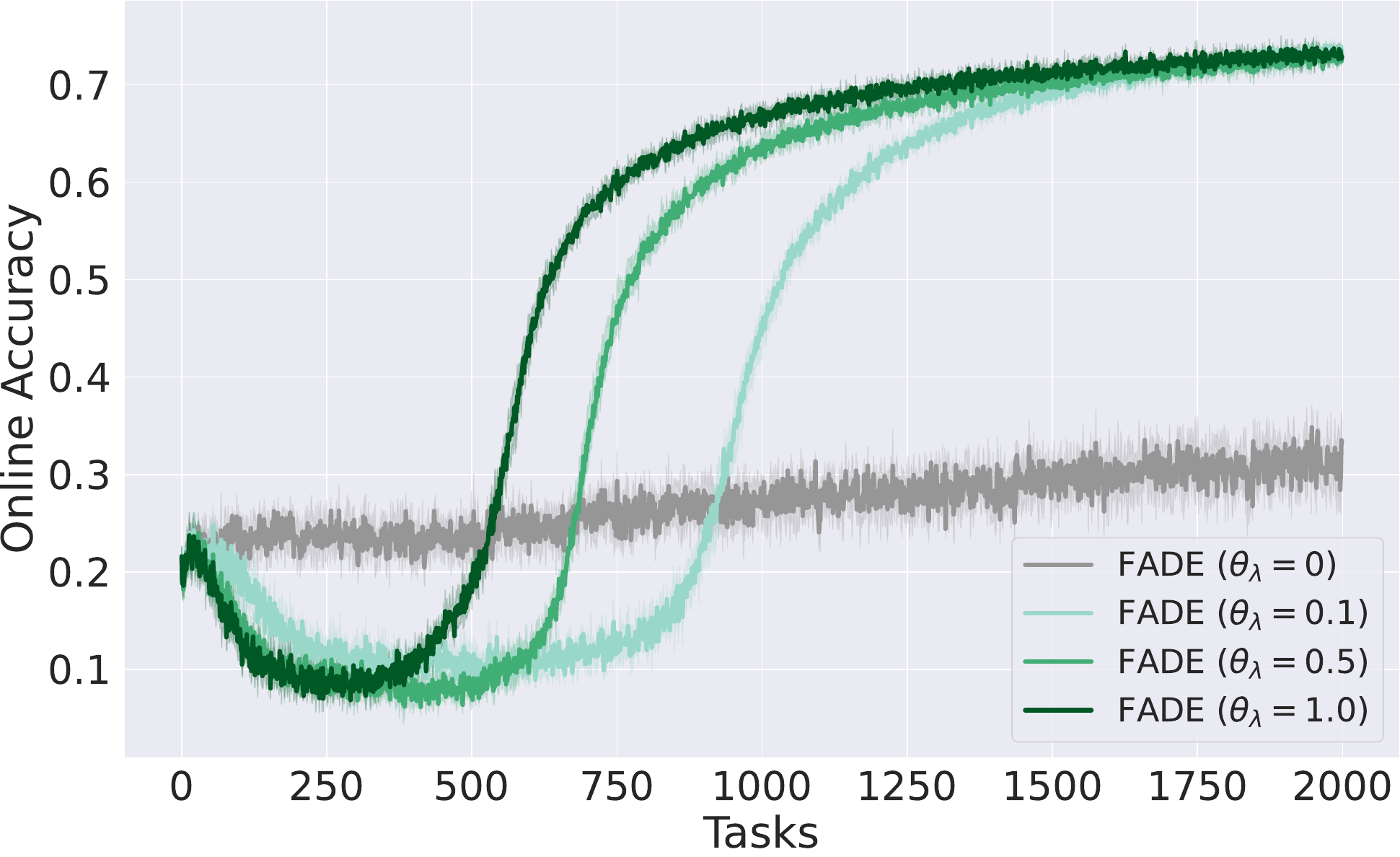}
        \caption{Online accuracy on label-permuted EMNIST with FADE + Adam, starting from $\gamma_0 = -11.5$ ($\lambda_0 \approx 10^{-5}$).
        Despite an initial drop, FADE with $\theta_{\lambda} > 0$ recovers strongly, with a final accuracy of over $0.7$.}
        \label{fig:emnist_adam_adaptive_learning_curves}
    \end{minipage}
\end{figure}

\paragraph{Results} 
Results are presented in Table~\ref{tab:emnist_base} and online accuracy across tasks is shown in Figure~\ref{fig:emnist_lp_full_learning_curves}.
Without weight decay, SGD and Adam perform poorly on this problem (with average accuracy of $0.258$ and $0.119$, respectively).
Adding weight decay to these approaches improves performance.
Weight clipping substantially improves performance, with SGD + WClip achieving an average online accuracy of $0.612$.
FADE outperforms the considered baselines, with FADE+SGD achieving an average accuracy of $0.807$ and FADE+Adam achieving $0.750$.

\begin{table}[b]
\centering
\caption{Average online accuracy across 5M interactions (2000 tasks) in the streaming classification problem with label permutation using EMNIST.}
\label{tab:emnist_base}
\begin{tabular}{lc}
\toprule
Method & Avg Online Accuracy \\
\midrule
SGD        & $0.258 \pm 0.006$ \\
SGD + WD  &  $0.335 \pm 0.002$ \\
Adam       & $0.119 \pm 0.003$ \\
AdamW     &  $0.372 \pm 0.001$ \\ 
SGD + WClip  & $0.612 \pm 0.003$  \\
\textbf{FADE + SGD}   & $\mathbf{0.807 \pm 0.001}$ \\
FADE + Adam  & $0.750\pm0.001$ \\
\bottomrule
\end{tabular}
\end{table}

\begin{table}[thb]
\centering
\caption{Impact of FADE's adaptive decay on label-permuted EMNIST with SGD.
Results are average online accuracy $\pm$ standard deviation across 5 seeds, with step size $\alpha=0.005$.
$\theta_\lambda=0$ corresponds to fixed weight decay on the head.}
\label{tab:emnist_adaptability_sgd_fade}
\begin{tabular}{lcccc}
\toprule
 & $\theta_\lambda=0$ & $\theta_\lambda=0.1$ &  $\theta_\lambda=0.5$ & $\theta_\lambda=1.0$\\
\midrule
$\gamma_0=-4.6\ (\lambda_0\approx0.01)$     & $0.694 \pm 0.001$ & $0.780\pm0.003$ & $0.798\pm0.001$ & $0.800\pm0.001$   \\
$\gamma_0=-6.9\ (\lambda_0\approx0.001)$    & $ 0.801\pm0.001$ & $\mathbf{0.807 \pm 0.001}$ & $0.806\pm0.001$ & $0.804 \pm 0.001$ \\
$\gamma_0=-9.2\ (\lambda_0\approx0.0001)$   &  $0.779\pm0.001$ & $0.801 \pm 0.001$  & $0.801\pm0.000$ & $0.798\pm0.001$ \\
$\gamma_0=-11.5\ (\lambda_0\approx0.00001)$ & $0.531 \pm 0.005$ & $0.682 \pm 0.003$  & $0.743 \pm 0.002$ & $0.750\pm 0.002$ \\
\bottomrule
\end{tabular}
\end{table}

\begin{table}[thb]
\centering
\caption{Impact of FADE's adaptive decay on label-permuted EMNIST with Adam.
Results are average online accuracy $\pm$ standard deviation across 5 seeds, with step size $\alpha=0.0001$.
$\theta_\lambda=0$ corresponds to fixed weight decay on the head.
}
\label{tab:emnist_adaptability_adam_fade}
\begin{tabular}{lcccc}
\toprule
 & $\theta_\lambda=0$ & $\theta_\lambda=0.1$ &  $\theta_\lambda=0.5$ & $\theta_\lambda=1.0$\\
\midrule
$\gamma_0=-4.6\ (\lambda_0\approx0.01)$     & $0.556\pm0.001$ & $0.737\pm0.000$ & $0.737\pm0.000$ & $0.719\pm0.001$   \\
$\gamma_0=-6.9\ (\lambda_0\approx0.001)$    & $0.746\pm0.001$ & $\mathbf{0.750\pm0.001}$ & $0.741 \pm 0.001$ & $0.721 \pm 0.001$ \\
$\gamma_0=-9.2\ (\lambda_0\approx0.0001)$   &  $0.588\pm0.001$ & $0.741\pm0.001$  & $0.716\pm0.002$ & $0.649\pm0.007$ \\
$\gamma_0=-11.5\ (\lambda_0\approx0.00001)$ & $0.269\pm0.004$ & $0.412\pm0.004$  & $0.472\pm0.005$ & $0.521\pm0.004$ \\
\bottomrule
\end{tabular}
\end{table}

We examine sensitivity to the initial decay rate $\lambda_0$ and meta-step-size $\theta_{\lambda}$ in Tables~\ref{tab:emnist_adaptability_sgd_fade} and~\ref{tab:emnist_adaptability_adam_fade}.
Notably, SGD (or Adam) with fixed weight decay applied only to the head can itself be a strong approach with the right decay coefficient (e.g., $0.801$ at $\gamma_0=-6.9$).
To our knowledge, such a baseline has not been explored in prior continual learning benchmarks.  
In label permutation, learned features remain useful across tasks, but the mapping from features to classes must be relearned after each permutation. 
Decay on the head could facilitate this relearning by clearing stale class mappings. 
This result highlights the importance of applying different decay rates to different parts of the network.
However, this baseline is fragile.
With fixed decay on the head $(\theta_\lambda=0)$, performance depends heavily on the choice of initial decay rate $\lambda_0$, ranging from $0.531$ to $0.801$ with SGD.
FADE reduces this sensitivity and provides substantial benefits when initialization is sub-optimal.
Figure~\ref{fig:emnist_adam_adaptive_learning_curves} illustrates the benefit of adaptivity from a poor initialization with FADE + Adam. 
With $\gamma_0 = -11.5 (\lambda_0 \approx 10^{-5})$, the initial decay is small and fixed decay performs poorly. FADE discovers appropriate decay rates during interaction, recovering to a high accuracy of over $0.7$.

\paragraph{FADE on all layers.} We evaluated a naive extension of FADE + SGD to all layers on the EMNIST benchmark, applying the same meta-gradient updates derived for the linear setting to hidden layers. 
The best configuration achieves
$0.535$ average online accuracy, outperforming SGD with fixed weight decay ($0.335$; Table~\ref{tab:emnist_base}) but well below head-only FADE ($0.807$; Table~\ref{tab:emnist_base}).
Furthermore, unlike head-only FADE, this variant is sensitive to initialization, indicating that the linear meta-gradient approximation is insufficient for hidden layers. 

\paragraph{Analysis with partial label permutations.}
We also evaluate a variant in which 24 of the 47 classes retain their labels across permutations, resulting in a mix of stable and changing outputs.
This setting may better reflect realistic continual learning problems where some structure persists across tasks.
The main results are presented in Table~\ref{tab:emnist_partial_perm}.
FADE+SGD achieves an average online accuracy of $0.841$ with the same hyperparameters as the variant where all classes were permuted, considerably better than the other baselines.
Further details and the sensitivity tables for FADE are provided in Appendix~\ref{app:extra_exps:partial_permuted_emnist}.
FADE’s benefits over its fixed counterpart are more pronounced in this setting.
The best fixed FADE+SGD (with $\theta_{\lambda}=0$) achieves $0.830$ vs $0.846$ for the best adaptive variant (see Table~\ref{tab:emnist_adaptability_sgd_fade_partial}).
Similarly, with FADE+Adam, the best $\theta_{\lambda}=0$ setting achieves $0.763$ vs $0.810$ for the best adaptive (see Table~\ref{tab:emnist_adaptability_adam_fade_partial}).

\begin{table}[t]
\centering
\caption{Average online accuracy across 5M interactions (2000 tasks) in the streaming classification problem with \emph{partial label permutation} using EMNIST.}
\label{tab:emnist_partial_perm}
\begin{tabular}{lc}
\toprule
Method & Avg Online Accuracy \\
\midrule
SGD &  $0.735 \pm 0.012$ \\
SGD + WD  &  $0.694 \pm 0.007 $ \\
Adam &  $0.448\pm0.016$ \\
AdamW     &  $0.541 \pm 0.004$ \\ 
SGD + WClip  & $0.719 \pm 0.006 $  \\
FADE + Adam  & $0.808 \pm 0.002$ \\
\textbf{FADE + SGD}   & $\mathbf{0.841 \pm 0.001}$ \\
\bottomrule
\end{tabular}
\end{table}

\section{Related Work}
\label{rwork}

Our approach is motivated by the perspective of learned forget gates applied to short-term memory in sequence-processing models~\citep{gers2000learning, beck2024xlstm, yang2024gated}.
The forget gate was introduced to prevent unbounded growth of memory cell states, enabling recurrent neural networks to discard outdated information stored in activations~\citep{gers2000learning, van2018unreasonable}.
Earlier, \citet{mozer1989focused} proposed an architecture that learns per-unit decay rates on the short-term memory via gradient descent with an online trace. 
FADE can be seen as applying these principles to long-term memories stored in weights, adapting per-parameter decay rates rather than per-activation gates.
Unlike activation-level forget gates, which are typically input-dependent, FADE's decay rates adapt slowly via meta-gradients, matching the timescale of the environment's non-stationarity rather than input-level variation.
In summary, FADE involves adapting weight decay via meta-learning for continual learning.

\paragraph{Adaptive weight decay.}
\citet{hanson1988comparing} study weight decay as a bias toward simpler networks, noting that uniform decay rates are suboptimal because all weights decay equally regardless of their importance. 
They compare two types of non-uniform weight decay at the hidden-unit level, with the general idea of applying stronger decay on small weights while leaving large weights intact.
In a Bayesian framework, \citet{mackay1992bayesian} proposes selecting weight decay coefficients via evidence maximization.
Several works have since explored adaptive weight decay for better regularization in problems with fixed, stationary datasets.
To balance the scale of loss gradients and the decay penalty, prior works have proposed adapting  weight decay coefficients across layers~\citep{ishii2017layer} or across training iterations to improve adversarial robustness~\citep{ghiasi2023improving}.
Taking adaptation to a more granular level, \citet{nakamura2019adaptive} propose per-parameter weight decay that scales with layer-wise normalized gradient magnitudes, effectively assigning stronger regularization to parameters with relatively larger gradients.
\citet{xie2023overlooked} propose a weight decay scheduler based on the gradient norm to mitigate the large gradient norms that hinder convergence and generalization when training deep neural networks.
Most recently, scaling adaptive weight decay to modern heterogeneous architectures, \citet{healphadecay} assign distinct weight decay strengths to different modules within large language models.
These methods are designed to improve generalization in mini-batch training with stationary targets.
FADE instead targets online continual learning, where non-stationarity rather than overfitting is the core challenge, and derives its adaptation from meta-gradient descent rather than gradient-norm heuristics.

\paragraph{Continual learning.} Agents with bounded capacity that are designed to learn forever must trade-off capacity for stability, or for plasticity, by either retaining what was learned against learning new things~\citep{grossberg1987competitive, elsayed2024addressing}.
With neural networks, continual learning can struggle on either component of this trade-off, leading to two distinct phenomena in the literature.
The first of these is 
catastrophic forgetting, where neural networks fail to retain what was previously learned when data changes over time~\citep{mccloskey1989catastrophic, ratcliff1990connectionist, french1999catastrophic}.
Mitigation strategies for catastrophic forgetting involve regularizing parameters towards past solutions \citep{kirkpatrick17_overc}.
It has also been recognized that storing historical examples and retraining provides an effective, but computationally demanding, baseline \citep{prabhu20_gdumb}.
More recently, loss of plasticity has been identified as another potential failure mode, where neural networks fail to adapt and learn new things~\citep{ash2020warm, dohare2024loss}.
This latter phenomenon is often mitigated by either regularization approaches~\citep{lewandowski2024learning, kumar2025maintaining}, or by re-initializing weights~\citep{nikishin2022primacy, hernandez2025reinitializing}.

\paragraph{Meta-learning for continual learning.}
The difficulties inherent to continual learning can broadly be addressed by meta-learning the learning algorithm itself, either through gradient-based meta-learning or program search~\citep{Schmidhuber:87long, Schmidhuber:92selfref, Schmidhuber:93selfrefann,
schmidhuber1997shifting,
hochreiter2001learning, andrychowicz2016learning, iriemetalearning}.
Unlike these more general approaches, FADE does not replace the entire learning algorithm.
It only adapts per-parameter decay rates, leaving the base optimizer (like SGD or Adam) intact.
Closely related are meta-gradient methods that adjust meta-parameters online via differentiation through the update rule~\citep{xu2018meta, zahavy2020self, luketina2022meta}.
IDBD~\citep{sutton1992adapting} uses meta-gradients to adapt per-parameter step sizes for online linear regression, and \citet{schraudolph1999local} extends it to nonlinear networks. \citet{sharifnassabmetaoptimize} generalize IDBD within a framework that optimizes meta-parameters against a discounted lifetime objective. 
FADE uses the same forward-mode meta-gradient derivation as IDBD, but focuses specifically on adapting weight decay rates, targeting how much to forget rather than how fast to learn.

\section{Conclusion}

We introduced FADE to adapt per-parameter weight decay rates online via forward-mode meta-gradients.
Our work was motivated by viewing weight decay as a mechanism for forgetting, rather than purely as a regularizer that controls weight magnitudes.
To enable judicious forgetting in continual learning, FADE automatically discovers distinct decay rates during interaction, adding minimal computational overhead.

Empirically, FADE consistently improves over fixed weight decay across three settings of increasing complexity. 
In a linear tracking problem, FADE complements step-size adaptation via IDBD, yielding the lowest error when combined. 
On a nonlinear teacher-student problem, applying FADE to the final layer achieves roughly half the error of AdamW.
On the streaming label-permuted EMNIST problem, FADE outperforms weight clipping, the previous best method.
Across all settings, FADE is robust to the initial decay rate, narrowing the performance gap across initializations considerably compared to fixed decay.

Our results reveal that applying a suitable fixed non-zero decay rate only to the final layer proves to be a strong baseline that has not received prior attention.
By default, weight decay applies a fixed decay rate uniformly across all parameters, making such a baseline easy to overlook. 
This finding supports the importance of different decay rates across different parts of the network, a principle that FADE automates at the per-parameter level on the final layer. 

We derived FADE for the online linear regression setting and applied it to the final layer when using it with neural networks.
Scaling FADE to larger architectures requires addressing some open challenges. 
We evaluated a naive extension of FADE to all layers on the EMNIST benchmark, applying the same meta-gradient updates derived for the linear setting to hidden layers as well. 
This variant outperforms SGD with fixed weight decay but plateaus below head-only FADE, suggesting that the linear approximation is insufficient for hidden layers.
Therefore, an important direction for future work is to formalize the interaction between the decay rates of the network head and those of the hidden layers, while accounting for nonlinearities between layers.
Another possibility is to design meta-gradient approximations that are invariant to the underlying architecture, making them easier to apply across different depths, nonlinearities, and layer types.
This would allow FADE to be extended to each layer in a network and combined with common architectural modifications, such as layer normalization, enabling its use in settings where non-stationarity arises naturally, such as reinforcement learning. 

\section*{Acknowledgments}
We thank Kazuki Irie, Vincent Herrmann, Arsalan Sharifnassab, and Nicolau Oliver for valuable discussions and helpful comments on earlier drafts.

\bibliography{collas2026_conference}
\bibliographystyle{collas2026_conference}

\appendix
\clearpage

\section{Derivations}
\label{app:derivations}

\subsection{FADE}
\label{app:derivations:fade_regression}

The weight update for $w_i$ is

\begin{equation}
\label{eq:fade_weight_update}
    w^i_{t+1} = (1 - \lambda^i_{t+1}) \, w^i_t + \alpha \, \delta_t \, x^i_t,
\end{equation} where $\delta_t = y^*_t - \sum_j w^j_t x^j_t$.

\paragraph{Deriving the $\gamma^i$ update} Per-parameter weight decay $\lambda^i = \exp({\gamma^i})$.

Stochastic gradient descent on loss $J_t =(\delta_t)^2/2$ with respect to $\gamma^i$:
\begin{equation*}
    \gamma^i_{t+1} = \gamma^i_t - \tfrac{1}{2}\theta_\lambda \frac{\partial (\delta_t)^2}{\partial \gamma_i}.
\end{equation*}

Expanding via the chain rule and applying the approximation from IDBD derivation - that the primary effect of changing a meta-parameter associated with weight $i$ is on weight $i$ itself, i.e., $\partial w^j_t / \partial \gamma^i \approx 0$ for $j \neq i$,
\begin{align*}
    \frac{\partial (\delta_t)^2}{\partial \gamma^i}
    &= \sum_j \frac{\partial (\delta_t)^2}{\partial w^j_t} \frac{\partial w^j_t}{\partial \gamma^i}
    \approx \frac{\partial (\delta_t)^2}{\partial w^i_t} \frac{\partial w^i_t}{\partial \gamma^i}.
\end{align*}

We have $\frac{\partial (\delta_t)^2}{\partial w^i_t} = -2\delta_t \, x^i_t$. Defining $g^i_t \triangleq \frac{\partial w^i_t}{\partial \gamma^i}$:
\begin{equation}
    \gamma^i_{t+1} = \gamma^i_t + \theta_\lambda \, \delta_t\,x^i_t \,g^i_t.
\end{equation}

\paragraph{Deriving the $g^i$ trace}

We differentiate (\ref{eq:fade_weight_update}) with respect to $\gamma^i$.
Note that $\frac{\partial \lambda^i}{\partial \gamma^i} = \lambda^i$ since $\lambda^i = \exp{\gamma^i}$.

\begin{align*}
    g^i_{t+1} &= \frac{\partial}{\partial \gamma^i}\Big[(1 - \lambda^i) w^i_t + \alpha \delta_t x^i_t\Big]\\
    &= -\lambda^i \, w^i_t + (1 - \lambda^i) \, g^i_t + \alpha \frac{\partial \delta_t}{\partial \gamma^i} x^i_t. \label{eq:app_g_expand}
\end{align*}

Next, $\frac{\partial \delta_t}{\partial \gamma^i} = \frac{\partial }{\partial \gamma^i} \left [- \sum_j w^j_t x^j_t \right ] \approx - g^i_t x^i_t$ using the same IDBD approximation as earlier.
This gives us

\begin{equation*}
    g^i_{t+1} = g^i_t\left (1 - \lambda^i - \alpha (x^i_t)^2\right ) - \lambda^i \, w^i_t.
\end{equation*}

Adding the positive-bounding operation for stability:

\begin{equation}
\label{eq:app_decoupled_g_final}
    g^i_{t+1} = g^i_t\big[1 - \lambda^i _{t+1} - \alpha (x^i_t)^2\big]^+ - \lambda^i_{t+1} \, w^i_t.
\end{equation}

\subsection{FADE + IDBD}
\label{app:derivations:fade_idbd}

Here, the per-parameter step size $\alpha^i = \exp{\beta^i}$.

The weight update for $w^i$ is

\begin{equation}
\label{eq:fade_idbd_weight_update}
    w^i_{t+1} = (1 - \lambda^i_{t+1}) \, w^i_t + \alpha^i_{t+1} \, \delta_t \, x^i_t,
\end{equation} where $\delta_t = y^*_t - \sum_j w^j_t x^j_t$.

The update for $\gamma$ and $g$ follows from the previous section.
We derive updates for $\beta$ and its associated trace $h$.
The original IDBD trace $h^i_t \triangleq \frac{\partial w^i_t}{\partial \beta^i}$ must be re-derived because of the introduction of weight decay.

Using the same approach as earlier,
\begin{equation}
    \beta^i_{t+1} = \beta^i_t + \theta_\alpha \, \delta_t \, x^i_t \, h^i_t.
\end{equation}

Differentiating Equation~\ref{eq:fade_idbd_weight_update} with respect to $\beta^i$, noting $\frac{\partial \alpha^i}{\partial \beta^i} = \alpha^i$ and $\lambda^i$ does not depend on $\beta^i$,

\begin{align*}
    h^i_{t+1} &= \frac{\partial}{\partial \beta^i}\Big[(1 - \lambda^i) w^i_t + \alpha^i \delta_t x^i_t\Big] \\
    &= (1 - \lambda^i) \, h^i_t + \alpha^i \, \delta_t \, x^i_t + \alpha^i \frac{\partial \delta_t}{\partial \beta^i} x^i_t.
\end{align*}

Using $\frac{\partial \delta_t}{\partial \beta^i} \approx -x^i_t h^i_t$ and collecting terms:
\begin{equation}
\label{eq:app_decoupled_h_final}
    h^i_{t+1} = h^i_t\big[1 - \lambda^i_{t+1} - \alpha^i_{t+1} (x^i_t)^2\big]^+ + \alpha^i_{t+1} \, \delta_t \, x^i_t.
\end{equation}
This is identical to IDBD's original $h^i$ update except that $\lambda^i$ appears in the decay factor.

\subsection{FADE with cross-entropy loss}
\label{app:derivations:fade_ce}

Let there be $C$ classes.
The learner produces logits $z^k_t = \sum_j W^{kj}_t\, x^j_t$ for $k = 1, \dots, C$,
where $x_t \in \mathbb{R}^d$ is the input, and weights $W_t \in \mathbb{R}^{C\times d}$.
The target $y_t \in \{1 \dots C\}$ is the class label.
The softmax probabilities are $p^k_t = \exp({z^k_t}) / \sum_{m} \exp({z^m_t})$,
and the cross-entropy loss for true class $y_t$ is
\begin{equation*}
    J_t = -\log p_t^{y_t}.
\end{equation*}

\paragraph{Gradient}
The gradient of $J$ with respect to the weights is
\begin{equation}
    \frac{\partial J_t}{\partial W^{kj}} = \bigl(p^k_t - \mathbb{I}[k = y_t]\bigr)\, x^j_t,
\end{equation}
where $\mathbb{I}[\cdot]$ is the indicator function.

\paragraph{Weight update}
As in the regression case, the weight update with per-parameter decay is
\begin{equation}
\label{eqn:fade_ce_weight_update}
    W^{kj}_{t+1} = (1 - \lambda^{kj}_{t+1})\, W^{kj}_t - \alpha \frac{\partial J_t}{\partial W^{kj}},
\end{equation}
where $\lambda^{kj} = \exp({\gamma^{kj}})$.

\paragraph{Meta-gradient update for $\gamma^{kj}$}
Following the same derivation as in Appendix~\ref{app:derivations:fade_regression}, 
\begin{equation*}
    \gamma^{kj}_{t+1} = \gamma^{kj}_t - \theta_\lambda \frac{\partial J_t}{\partial \gamma^{kj}}.
\end{equation*}

Expanding via the chain rule and applying the same approximation 
as in the regression case --- that the primary effect of changing 
$\gamma^{kj}$ is on $W^{kj}$ itself, i.e., 
$\partial W^{k'j'}_t / \partial \gamma^{kj} \approx 0$ for $(k',j') \neq (k,j)$,
\begin{align*}
    \frac{\partial J_t}{\partial \gamma^{kj}}
    &= \sum_{k',j'} \frac{\partial J_t}{\partial W^{k'j'}_t} 
    \frac{\partial W^{k'j'}_t}{\partial \gamma^{kj}}
    \approx \frac{\partial J_t}{\partial W^{kj}_t} 
    \frac{\partial W^{kj}_t}{\partial \gamma^{kj}}.
\end{align*}

Defining $g^{kj}_t \triangleq \partial W^{kj}_t / \partial \gamma^{kj}$ and $\delta^k_t \triangleq \bigl(\mathbb{I}[k = y_t] - p^k_t \bigr)$:

\begin{equation}
    \gamma^{kj}_{t+1} = \gamma^{kj}_t + \theta_\lambda \, \delta^k_t \,x^j_t \, g^{kj}_t
\end{equation}
So far the derivation is analogous to the regression case, the meta-gradient uses the 
loss gradient $-\partial J / \partial W^{kj}$ multiplied by the trace $g^{kj}$.

\paragraph{Trace update}
We differentiate Equation~\ref{eqn:fade_ce_weight_update} with respect to $\gamma^{kj}$

\begin{align*}
    g^{kj}_{t+1} &= \frac{\partial}{\partial \gamma^{kj}}
    \Big[(1 - \lambda^{kj}) W^{kj}_t - \alpha (p^k_t - \mathbb{I}[k=y_t]) x^j_t\Big] \\
    &= -\lambda^{kj} \, W^{kj}_t + (1 - \lambda^{kj}) \, g^{kj}_t 
    - \alpha \frac{\partial p^k_t}{\partial \gamma^{kj}} x^j_t.
\end{align*}

For the remaining term, we expand using the chain rule through the logits:
\begin{equation*}
    \frac{\partial p^k_t}{\partial \gamma^{kj}} 
    = \sum_{k'} \frac{\partial p^k}{\partial z^{k'}} 
    \cdot \frac{\partial z^{k'}}{\partial \gamma^{kj}}.
\end{equation*}
Applying the approximation, 
$\partial W^{k'j'} / \partial \gamma^{kj} \approx 0$ for $(k',j') \neq (k,j)$, 
only the $k' = k$ term survives:
\begin{equation*}
    \frac{\partial z^{k'}}{\partial \gamma^{kj}} 
    = \sum_{j'} x^{j'} \frac{\partial W^{k'j'}}{\partial \gamma^{kj}} 
    \approx \begin{cases} x^j \, g^{kj}_t & \text{if } k' = k \\ 0 & \text{otherwise} \end{cases}
\end{equation*}
Using the standard softmax derivative 
$\partial p^k / \partial z^k = p^k(1 - p^k)$:
\begin{equation*}
    \frac{\partial p^k_t}{\partial \gamma^{kj}} 
    \approx p^k_t(1 - p^k_t) \cdot x^j_t \cdot g^{kj}_t.
\end{equation*}
Substituting:
\begin{equation*}
    g^{kj}_{t+1} = g^{kj}_t\Big[1 - \lambda^{kj} 
    - \alpha\, p^k_t(1 - p^k_t)\, (x^j_t)^2\Big] 
    - \lambda^{kj}\, W^{kj}_t.
\end{equation*}
Adding the positive-bounding operation for stability:
\begin{equation}
    g^{kj}_{t+1} = g^{kj}_t\Big[1 - \lambda^{kj}_{t+1}
    - \alpha\, p^k_t(1 - p^k_t)\, (x^j_t)^2\Big]^+ 
    - \lambda^{kj}_{t+1}\, W^{kj}_t.
\end{equation}
Compared to the regression trace (Equation~\ref{eq:app_decoupled_g_final}), 
the only difference is the factor $p^k_t(1 - p^k_t) \in [0, 0.25]$ 
multiplying $(x^j_t)^2$.

\subsection{FADE with Adam}
\label{app:derivations:fade_adam}

When combining FADE with Adam, the base optimizer provides an effective per-parameter step size. 

At step $t$, Adam's effective step size for parameter $i$ is
\begin{equation}
\alpha_{\text{eff},t}^i = \frac{\alpha}{\sqrt{\hat{v}_t^i} + \epsilon},
\end{equation}
where $\hat{v}_t^i = v_t^i / (1 - \beta_2^t)$ is the bias-corrected second moment estimate~\citep{kingma2014adam}.
The FADE trace update (Equation~\ref{eq:app_decoupled_g_final}) becomes
\begin{equation}
g_{t+1}^i \leftarrow g_t^i \left[1 - \lambda_{t+1}^i - \alpha_{\text{eff},t}^i (x_t^i)^2\right]_+ - \lambda_{t+1}^i w_t^i.
\end{equation}

The $\gamma$ update and weight decay application remain unchanged. 

Adam handles the gradient step, and FADE applies adaptive weight decay independently, analogous to the relationship between Adam and AdamW.
Note that this approximation does not account for the effect of Adam's first moment (momentum) on the trace, which would require tracking additional dependencies.

\subsection{Coupled decay and step size adaptation for online linear regression}
\label{app:derivations:coupled}

Here we explore an alternative to FADE+IDBD where the weight update comes from SGD on the $L_2$ regularized loss, i.e.

\begin{equation}
\label{eq:coupled_fade_update}
    w^i_{t+1} = (1 - \alpha^i_{t+1}\lambda^i_{t+1}) \, w^i_t + \alpha^i_{t+1} \, \delta_t \, x^i_t.
\end{equation}

Note that in this update, the decay term involves a product with the per-parameter step size.
We proceed with the same derivation technique.

\paragraph{Updates to $\gamma^i$ and $\beta^i$}
The meta-parameter updates follow the same form as in the decoupled case:
\begin{align}
    \beta^i_{t+1} &= \beta^i_t + \theta_\alpha \, \delta_t \, x^i_t \, h^i_t, \quad \alpha^i_{t+1} = \exp(\beta^i_{t+1}), \\
    \gamma^i_{t+1} &= \gamma^i_t + \theta_\lambda \, \delta_t \, x^i_t \, g^i_t, \quad \lambda^i_{t+1} = \exp(\gamma^i_{t+1}).
\end{align}

The traces $g^i_t \triangleq \partial w^i_t / \partial \gamma^i$ and $h^i_t \triangleq \partial w^i_t / \partial \beta^i$ must be re-derived to account for the coupling between $\alpha^i$ and $\lambda^i$ in the decay term.

\paragraph{Deriving the $g^i$ trace}
We differentiate the weight update~(\ref{eq:coupled_fade_update}) with respect to $\gamma^i$.
Note that $\partial \lambda^i / \partial \gamma^i = \lambda^i$ and $\alpha^i$ does not depend on $\gamma^i$:
\begin{align*}
    g^i_{t+1} &= \frac{\partial}{\partial \gamma^i}\Big[(1 - \alpha^i \lambda^i) w^i_t + \alpha^i \delta_t x^i_t\Big] \\
    &= -\alpha^i \lambda^i \, w^i_t + (1 - \alpha^i \lambda^i) \, g^i_t + \alpha^i \frac{\partial \delta_t}{\partial \gamma^i} x^i_t.
\end{align*}
Using $\partial \delta_t / \partial \gamma^i \approx -g^i_t x^i_t$ and adding the positive-bounding operation:
\begin{equation}
    g^i_{t+1} = g^i_t \big[1 - \alpha^i_{t+1} \lambda^i_{t+1} - \alpha^i_{t+1} (x^i_t)^2\big]^+ - \alpha^i_{t+1} \lambda^i_{t+1} \, w^i_t.
\end{equation}
This has a similar structure to the decoupled case (Equation~\ref{eq:app_decoupled_g_final}), but with $\lambda^i$ replaced by $\alpha^i \lambda^i$ throughout.

\paragraph{Deriving the $h^i$ trace}
We differentiate the weight update~(\ref{eq:coupled_fade_update}) with respect to $\beta^i$.
Note that $\partial \alpha^i / \partial \beta^i = \alpha^i$. Crucially, because the decay term is $\alpha^i \lambda^i$, it now depends on $\beta^i$:
\begin{align*}
    h^i_{t+1} &= \frac{\partial}{\partial \beta^i}\Big[(1 - \alpha^i \lambda^i) w^i_t + \alpha^i \delta_t x^i_t\Big] \\
    &= -\alpha^i \lambda^i \, w^i_t + (1 - \alpha^i \lambda^i) \, h^i_t + \alpha^i \delta_t \, x^i_t + \alpha^i \frac{\partial \delta_t}{\partial \beta^i} x^i_t.
\end{align*}
Using $\partial \delta_t / \partial \beta^i \approx -h^i_t x^i_t$ and adding the positive-bounding operation:
\begin{equation}
\label{eq:coupled_h_trace}
    h^i_{t+1} = h^i_t \big[1 - \alpha^i_{t+1} \lambda^i_{t+1} - \alpha^i_{t+1} (x^i_t)^2\big]^+ + \alpha^i_{t+1} \delta_t \, x^i_t - \alpha^i_{t+1} \lambda^i_{t+1} \, w^i_t.
\end{equation}
Compared to the decoupled $h^i$ trace (Equation~\ref{eq:app_decoupled_h_final}), which has the form
\begin{equation*}
    h^i_{t+1} = h^i_t \big[1 - \lambda^i_{t+1} - \alpha^i_{t+1} (x^i_t)^2\big]^+ + \alpha^i_{t+1} \delta_t \, x^i_t,
\end{equation*}
the coupled trace acquires an additional $-\alpha^i_{t+1} \lambda^i_{t+1} \, w^i_t$ term.

\section{Implementation details}
\label{app:impdets}

\subsection{Linear tracking}
\label{app:impdets:linear_tracking}

\paragraph{Hyperparameter search}
For each algorithm, we perform a grid search over step size and meta-parameters, evaluating MSE averaged over all 200K steps across 10 seeds.
\textbf{Selected (best)} hyperparameters are indicated in \textbf{bold}.

\subsubsection{Noise $\sigma_n = 0$}

\paragraph{SGD} Step size $\alpha \in \{0.5, 0.1, \mathbf{0.05}, 0.01, 0.005, 0.001\}$.
No weight decay, i.e., $\lambda=0$.

\paragraph{SGD + WD} Step size $\alpha \in \{0.5, 0.1, \mathbf{0.05}, 0.01, 0.005, 0.001\}$, weight decay $\lambda \in \{0.001, 0.005, 0.01, \mathbf{0.05}, 0.1, 0.5\}$.

\paragraph{IDBD} Meta step size $\theta_{\alpha} \in \{0.5, 0.1, 0.05, 0.01, \mathbf{0.005}, 0.001\}$, initial step size $\beta_0 \in \{-0.7, -2.3, -3, \mathbf{-4.6}, -5.3\}$.

\paragraph{IDBD + WD} Meta step size $\theta_{\alpha} \in \{0.1, 0.05, \mathbf{0.01}, 0.005\}$, initial step size $\beta_0 \in \{-0.7, -2.3, -3, \mathbf{-4.6}, -5.3\}$, weight decay $\lambda \in \{0.001, 0.005, \mathbf{0.01}, 0.05, 0.1, 0.5\}$.

\paragraph{FADE} Step size $\alpha \in \{\mathbf{0.1}, 0.05, 0.01, 0.005\}$, meta step size $\theta_\lambda \in \{0.1, 0.05, \mathbf{0.01}, 0.005\}$, initial decay $\gamma_0 \in \{0, -0.7, \mathbf{-1.2}, -2.3, -4.6, -6.9, -9.2\}$.

\paragraph{FADE + IDBD} Tied meta step size $\theta_\alpha = \theta_\lambda \in \{0.1, 0.05, \mathbf{0.01}, 0.005\}$, initial step size $\beta_0 \in \{-0.7, -2.3, -3, \mathbf{-4.6}, -5.3\}$, initial decay $\gamma_0 \in \{-0.7, -1.2, \mathbf{-2.3}, -4.6\}$.

\paragraph{Coupled WD + SS Adaptation} Tied meta step size $\theta_\alpha = \theta_\lambda \in \{0.1, 0.05, 0.01, 0.005\}$, initial step size $\beta_0 \in \{-0.7, -2.3, -3, -4.6, -5.3\}$, initial decay $\gamma_0 \in \{-0.7, -1.2, -2.3, -4.6\}$.

\subsubsection{Noise $\sigma_n = 1.0$}

\paragraph{SGD} Step size $\alpha \in \{0.5, 0.1, 0.05, \mathbf{0.01}, 0.005, 0.001\}$.
No weight decay, i.e., $\lambda=0$.

\paragraph{SGD + WD} Step size $\alpha \in \{0.5, 0.1, \mathbf{0.05}, 0.01, 0.005, 0.001\}$, weight decay $\lambda \in \{0.001, 0.005, 0.01, \mathbf{0.05}, 0.1, 0.5\}$.

\paragraph{IDBD} Meta step size $\theta_{\alpha} \in \{0.5, 0.1, 0.05, \mathbf{0.01}, 0.005, 0.001\}$, initial step size $\beta_0 \in \{-0.7, -2.3, -3, -4.6, \mathbf{-5.3}, -6.9\}$.
No weight decay, i.e., $\lambda=0$.

\paragraph{IDBD + WD} Meta step size $\theta_\alpha \in \{0.1, 0.05, \mathbf{0.01}, 0.005\}$, initial step size $\beta_0 \in \{-0.7, -2.3, -3, \mathbf{-4.6}, -5.3\}$, weight decay $\lambda \in \{0.001, 0.005, \mathbf{0.01}, 0.05, 0.1, 0.5\}$.

\paragraph{FADE} Step size $\alpha \in \{0.1, \mathbf{0.05}, 0.01, 0.005\}$, meta step size $\theta_\lambda \in \{0.1, 0.05, \mathbf{0.01}, 0.005\}$, initial decay $\gamma_0 \in \{0, -0.7, \mathbf{-1.2}, -2.3, -4.6, -6.9, -9.2\}$. 

\paragraph{FADE + IDBD} Tied meta step size $\theta_\alpha = \theta_\lambda \in \{0.1, 0.05, \mathbf{0.01}, 0.005\}$, initial step size $\beta_0 \in \{-0.7, -2.3, -3, \mathbf{-4.6}, -5.3\}$, initial decay $\gamma_0 \in \{-0.7, -1.2, \mathbf{-2.3}, -4.6\}$.

\paragraph{Coupled WD + SS Adaptation} Tied meta step size $\theta_\alpha = \theta_\lambda \in \{0.1, 0.05, 0.01, 0.005\}$, initial step size $\beta_0 \in \{-0.7, -2.3, -3, -4.6, -5.3\}$, initial decay $\gamma_0 \in \{-0.7, -1.2, -2.3, -4.6\}$.

\subsection{Non-linear tracking}
\label{app:impdets:nonlinear_tracking}

\paragraph{Hyperparameter search}
For each algorithm, we perform a grid search over step size and regularization parameters, evaluating MSE averaged over the final 500K of 2M training steps across 5 seeds.
All other unspecified hyperparameters (e.g., $\beta_1, \beta_2, \epsilon$ for Adam~\citep{kingma2014adam}) are PyTorch defaults.
\textbf{Selected (best)} hyperparameters are indicated in \textbf{bold}.

\paragraph{SGD} Step size $\alpha \in \{0.1, \mathbf{0.01}, 0.001, 0.0001\}$. No weight decay, i.e., $\lambda=0$.

\paragraph{Adam} Step size $\alpha \in \{0.1, 0.01, \mathbf{0.001}, 0.0001\}$. 
No weight decay, i.e., $\lambda=0$.

\paragraph{SGD + WD} Step size $\alpha \in \{\ 0.1, \mathbf{0.01}, 0.001, 0.0001\}$ and weight decay $\lambda \in \{1e-5, 1e-4, \mathbf{1e-3}, 1e-2, 1e-1 \}$.

\paragraph{AdamW} Step size $\alpha \in \{0.1, 0.01, \mathbf{0.001}, 0.0001\}$. 
Weight decay $\lambda \in \{10, 1, \mathbf{0.1}, 0.01, 0.001, 0.0001\}$.

\paragraph{FADE + SGD} Step size $\alpha \in \{0.1, \mathbf{0.01}, 0.001\}$, meta step size $\theta_{\lambda} \in \{0, 0.5, 1 ,\mathbf{2}\}$.
We also consider (and report results for) different values for $\gamma_0 \in \{-2.3, -4.6, -6.9, \mathbf{-9.2} \}$.

\paragraph{FADE + Adam} Step size $\alpha \in \{0.01, 0.001, \mathbf{0.0001}, 0.00001\}$, meta step size $\theta_{\lambda} \in \{0, 0.5, \mathbf{1} ,2\}$.
We also consider different values for $\gamma_0 \in \{-2.3, -4.6, -6.9, \mathbf{-9.2} \}$.

\subsection{Streaming image classification with label permutation}
\label{app:impdets:label_perm}

For each algorithm, we perform a grid search over step size and regularization parameters, evaluating online accuracy averaged over all 5M training steps across 5 seeds.
All other unspecified hyperparameters (e.g., $\beta_1, \beta_2, \epsilon$ for Adam~\citep{kingma2014adam}) are PyTorch defaults.
\textbf{Selected (best)} hyperparameters are indicated in \textbf{bold}.

\paragraph{SGD.} Step size $\alpha \in \{0.05, \mathbf{0.01}, 0.005, 0.001, 0.0005\}$.

\paragraph{SGD + WD.} Step size $\alpha \in \{0.05, \mathbf{0.01}, 0.005, 0.001\}$, and weight decay $\lambda \in \{1e-2, 1e-3, \mathbf{1e-4}, 1e-5, 1e-6 \}$.

\paragraph{Adam.} Step size $\alpha \in \{0.005, 0.001, \mathbf{0.0005}, 0.0001, 0.00005\}$.

\paragraph{AdamW.} Step size $\alpha \in \{0.005, 0.001, \mathbf{0.0005}, 0.0001, 0.00005\}$, and weight decay $\lambda \in \{1, \mathbf{0.1}, 0.01, 0.001, 0.0001\}$.

\paragraph{SGD + Weight Clipping.} Step size $\alpha \in \{0.1, 0.05, 0.01, \mathbf{0.005}, 0.001\}$, 
clipping parameter $\kappa \in \{1, \mathbf{2}, 3, 4, 5\}$. 

\paragraph{FADE + SGD.} Step size $\alpha \in \{0.05, 0.01, \mathbf{0.005}, 0.001\}$,
meta-step size $\theta_{\lambda} \in \{0, \mathbf{0.1}, 0.5,1 \}$, and $\gamma_0 \in \{-2.3, -4.6, \mathbf{-6.9}, -9.2 \}$

\paragraph{FADE + Adam.} Step size $\alpha \in \{0.001, 0.0005, \mathbf{0.0001}, 0.00005\}$,
meta-step size $\theta_{\lambda} \in \{0, \mathbf{0.1}, 0.5,1 \}$, and $\gamma_0 \in \{-2.3, -4.6, \mathbf{-6.9}, -9.2\}$

\section{Streaming classification with partial label permuted EMNIST}
\label{app:extra_exps:partial_permuted_emnist}

\begin{figure}[t]
\centering
\includegraphics[width=0.6\linewidth]{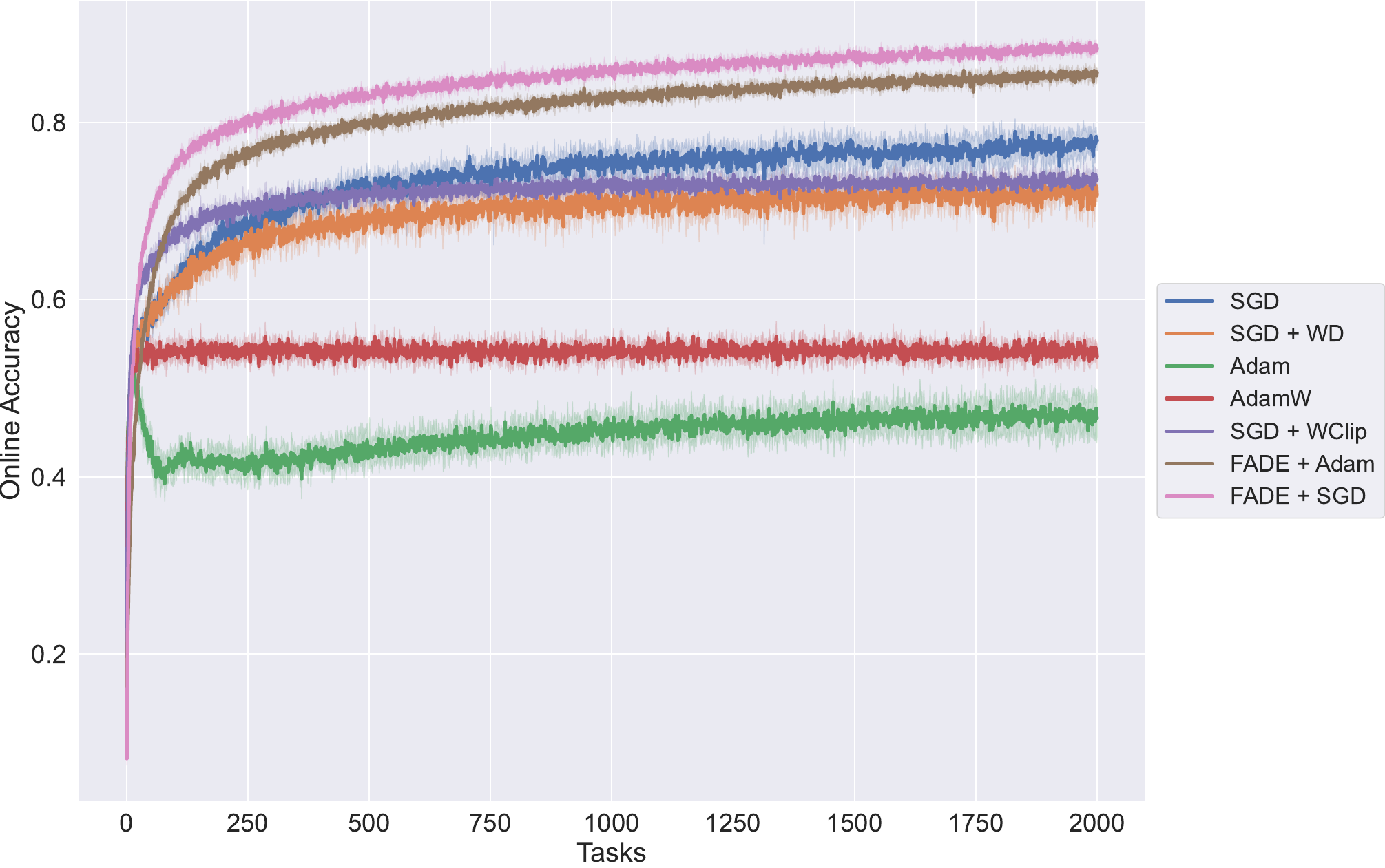}
\caption{Online accuracy with partial label permutations.}
\label{fig:emnist_lp_partial_learning_curves}
\end{figure}

We use the same selected hyperparameters as the setting with label permutations applied to all classes,
these results are provided in Table~\ref{tab:emnist_partial_perm} and Figure~\ref{fig:emnist_lp_partial_learning_curves}.

In this setting, SGD without weight decay outperforms SGD with weight decay
(Table~\ref{tab:emnist_partial_perm}).
This further motivates weight-decay adaptation based on the problem.

Sensitivity analysis for FADE + SGD and FADE + Adam are provided in Table~\ref{tab:emnist_adaptability_sgd_fade_partial} and Table~\ref{tab:emnist_adaptability_adam_fade_partial}.
Some alternative hyperparameter settings for FADE are slightly better than the default settings used from the setup where all labels were permuted (e.g. $0.841$ vs $0.846$ in Table~\ref{tab:emnist_adaptability_sgd_fade_partial}).

\begin{table}[thb]
\centering
\caption{Impact of FADE's adaptive decay on \emph{partial label-permuted} EMNIST (24 stable classes which never change labels) with SGD.
Results are average online accuracy $\pm$ standard deviation across 5 seeds, with step size $\alpha=0.005$.
$\theta_\lambda=0$ corresponds to fixed weight decay on the head.}
\label{tab:emnist_adaptability_sgd_fade_partial}
\begin{tabular}{lcccc}
\toprule
 & $\theta_\lambda=0$ & $\theta_\lambda=0.1$ &  $\theta_\lambda=0.5$ & $\theta_\lambda=1.0$\\
\midrule
$\gamma_0=-4.6\ (\lambda_0\approx0.01)$     & $0.699 \pm 0.000$ & $0.811 \pm 0.005$ & $0.816\pm0.006$ & $0.822\pm0.006$   \\
$\gamma_0=-6.9\ (\lambda_0\approx0.001)$    & $0.810\pm0.001$ & $0.841\pm0.001$ & $0.845\pm0.002$ & $0.846\pm0.003$ \\
$\gamma_0=-9.2\ (\lambda_0\approx0.0001)$   &  $0.830\pm0.002$ & $0.845\pm0.002$  & $0.846\pm0.002$ & $0.845\pm0.002$ \\
$\gamma_0=-11.5\ (\lambda_0\approx0.00001)$ & $0.783\pm0.006$ & $0.818\pm0.003$  & $0.833\pm0.003$ & $0.835\pm0.002$ \\
\bottomrule
\end{tabular}
\end{table}

\begin{table}[t]
\centering
\caption{Impact of FADE's adaptive decay on \emph{partial label-permuted} EMNIST (24 stable classes which never change labels) with Adam.
Results are average online accuracy $\pm$ standard deviation across 5 seeds, with step size $\alpha=0.0001$.
$\theta_\lambda=0$ corresponds to fixed weight decay on the head.}
\label{tab:emnist_adaptability_adam_fade_partial}
\begin{tabular}{lcccc}
\toprule
 & $\theta_\lambda=0$ & $\theta_\lambda=0.1$ &  $\theta_\lambda=0.5$ & $\theta_\lambda=1.0$\\
\midrule
$\gamma_0=-4.6\ (\lambda_0\approx0.01)$     & $0.560\pm0.001$ & $0.799\pm0.001$ & $0.807\pm0.002$ & $0.800\pm0.002$   \\
$\gamma_0=-6.9\ (\lambda_0\approx0.001)$    & $0.763\pm0.000$ & $0.808\pm0.002$ & $0.808\pm0.002$ & $0.803\pm0.003$ \\
$\gamma_0=-9.2\ (\lambda_0\approx0.0001)$   &  $0.753\pm0.002$ & $0.810\pm0.001$  & $0.810\pm0.002$ & $0.797\pm0.002$ \\
$\gamma_0=-11.5\ (\lambda_0\approx0.00001)$ & $0.549\pm0.011$ & $0.728\pm0.005$  & $0.732\pm0.003$ & $0.734\pm0.005$ \\
\bottomrule
\end{tabular}
\end{table}

\section{Algorithms}
We present pseudocode for IDBD (Algorithm~\ref{alg:idbd}), IDBD with fixed weight decay (Algorithm~\ref{alg:idbd-wd}), and FADE + IDBD (Algorithm~\ref{alg:fade-idbd}).
These algorithms are used in the linear tracking experiment (Section~\ref{exps:linear}). 
IDBD + WD is a special case of FADE + IDBD in which the decay rate $\lambda$
is fixed rather than adapted.
Removing the $\gamma$ update and 
$g$ trace from Algorithm~\ref{alg:fade-idbd} and replacing the adaptive $\lambda^i_{t+1}$
with a fixed scalar $\lambda$ recovers Algorithm~\ref{alg:idbd-wd}.
We tie meta-step sizes $\theta_{\alpha} = \theta_{\lambda} = \theta$ in our experiment with FADE+IDBD.

\begin{algorithm}[tbh]
\caption{IDBD: Incremental Delta-Bar-Delta \citep{sutton1992adapting}}
\label{alg:idbd}
\begin{algorithmic}[1]
\REQUIRE meta-step size $\theta_\alpha$, initial step-size parameter $\beta_0 \in \mathbb{R}^d$
\STATE Initialize weights $w_0 \in \mathbb{R}^d$, traces $h_0 \leftarrow \mathbf{0} \in \mathbb{R}^d$
\STATE Initialize $\alpha_0^i \leftarrow \exp({\beta_0^i})$ for all $i$
\FOR{$t = 0, 1, 2, \ldots$}
    \STATE Receive input $x_t \in \mathbb{R}^d$ and target $y^*_t \in \mathbb{R}$
    \STATE Predict $y_t \leftarrow \langle w_t, x_t \rangle$
    \STATE Compute error $\delta_t \leftarrow y^*_t - y_t$
    \FOR{each parameter $i = 1, \ldots, d$}
        \STATE \textit{\# Adapt step size}
        \STATE $\beta_{t+1}^i \leftarrow \beta_t^i + \theta_\alpha \, \delta_t \, x_t^i \, h_t^i$
        \STATE $\alpha_{t+1}^i \leftarrow \exp({\beta_{t+1}^i})$
        \STATE \textit{\# Update weight}
        \STATE $w_{t+1}^i \leftarrow w_t^i + \alpha_{t+1}^i \, \delta_t \, x_t^i$
        \STATE \textit{\# Update sensitivity trace}
        \STATE $h_{t+1}^i \leftarrow h_t^i \big[1 - \alpha_{t+1}^i (x_t^i)^2\big]^{+} + \alpha_{t+1}^i \, \delta_t \, x_t^i$
    \ENDFOR
\ENDFOR
\end{algorithmic}
\end{algorithm}

\begin{algorithm}[bt]
\caption{IDBD + WD: Adaptive Step Size with Fixed Weight Decay (online linear regression)}
\label{alg:idbd-wd}
\begin{algorithmic}[1]
\REQUIRE meta-step size $\theta_\alpha$, initial step-size parameter $\beta_0 \in \mathbb{R}^d$, weight decay $\lambda \in \mathbb{R}$
\STATE Initialize weights $w_0 \in \mathbb{R}^d$, traces $h_0 \leftarrow \mathbf{0} \in \mathbb{R}^d$
\STATE Initialize $\alpha_0^i \leftarrow \exp({\beta_0^i})$ for all $i$
\FOR{$t = 0, 1, 2, \ldots$}
    \STATE Receive input $x_t \in \mathbb{R}^d$ and target $y^*_t \in \mathbb{R}$
    \STATE Predict $y_t \leftarrow \langle w_t, x_t \rangle$
    \STATE Compute error $\delta_t \leftarrow y^*_t - y_t$
    \FOR{each parameter $i = 1, \ldots, d$}
        \STATE \textit{\# Adapt step size}
        \STATE $\beta_{t+1}^i \leftarrow \beta_t^i + \theta_\alpha \, \delta_t \, x_t^i \, h_t^i$
        \STATE $\alpha_{t+1}^i \leftarrow \exp({\beta_{t+1}^i})$
        \STATE \textit{\# Update weight with fixed decay}
        \STATE $w_{t+1}^i \leftarrow (1 - \lambda) \, w_t^i + \alpha_{t+1}^i \, \delta_t \, x_t^i$
        \STATE \textit{\# Update sensitivity trace}
        \STATE $h_{t+1}^i \leftarrow h_t^i \big[1 - \lambda - \alpha_{t+1}^i (x_t^i)^2\big]^{+} + \alpha_{t+1}^i \, \delta_t \, x_t^i$
    \ENDFOR
\ENDFOR
\end{algorithmic}
\end{algorithm}

\begin{algorithm}[t]
\caption{FADE + IDBD: Adaptive Decay and Step Size (online linear regression)}
\label{alg:fade-idbd}
\begin{algorithmic}[1]
\REQUIRE meta-step sizes $\theta_\alpha, \theta_\lambda$ (can be tied: $\theta_\alpha = \theta_\lambda = \theta$), initial $\beta_0, \gamma_0 \in \mathbb{R}^d$
\STATE Initialize weights $w_0 \in \mathbb{R}^d$, traces $h_0 \leftarrow \mathbf{0}$, $g_0 \leftarrow \mathbf{0} \in \mathbb{R}^d$
\STATE Initialize $\alpha_0^i \leftarrow \exp({\beta_0^i})$, $\lambda_0^i \leftarrow \exp({\gamma_0^i})$ for all $i$
\FOR{$t = 0, 1, 2, \ldots$}
    \STATE Receive input $x_t \in \mathbb{R}^d$ and target $y^*_t \in \mathbb{R}$
    \STATE Predict $y_t \leftarrow \langle w_t, x_t \rangle$, compute error $\delta_t \leftarrow y^*_t - y_t$
    \FOR{each parameter $i = 1, \ldots, d$}
        \STATE \textit{\# Adapt step size and decay rate}
        \STATE $\beta_{t+1}^i \leftarrow \beta_t^i + \theta_\alpha \, \delta_t \, x_t^i \, h_t^i$, \quad $\alpha_{t+1}^i \leftarrow \exp({\beta_{t+1}^i})$
        \STATE $\gamma_{t+1}^i \leftarrow \gamma_t^i + \theta_\lambda \, \delta_t \, x_t^i \, g_t^i$, \quad $\lambda_{t+1}^i \leftarrow \exp({\gamma_{t+1}^i})$
        \STATE \textit{\# Update traces}
        \STATE $h_{t+1}^i \leftarrow h_t^i \big[1 - \lambda_{t+1}^i - \alpha_{t+1}^i (x_t^i)^2\big]^{+} + \alpha_{t+1}^i \, \delta_t \, x_t^i$
        \STATE $g_{t+1}^i \leftarrow g_t^i \big[1 - \lambda_{t+1}^i - \alpha_{t+1}^i (x_t^i)^2\big]^{+} - \lambda_{t+1}^i \, w_t^i$
        \STATE \textit{\# Update weight}
        \STATE $w_{t+1}^i \leftarrow (1 - \lambda_{t+1}^i) \, w_t^i + \alpha_{t+1}^i \, \delta_t \, x_t^i$
    \ENDFOR
\ENDFOR
\end{algorithmic}
\end{algorithm}

\end{document}

%% file: math_commands.tex
\usepackage{amsmath,amsfonts,bm}

\def\eqref#1{equation~\ref{#1}}

\def\1{\bm{1}}

\DeclareMathAlphabet{\mathsfit}{\encodingdefault}{\sfdefault}{m}{sl}
\SetMathAlphabet{\mathsfit}{bold}{\encodingdefault}{\sfdefault}{bx}{n}